\documentclass{article}
\usepackage{arxiv}
\pdfoutput=1
\DeclareUnicodeCharacter{00A0}{ }
\DeclareUnicodeCharacter{0301}{\'{e}}
\usepackage[T1]{fontenc}    % use 8-bit T1 fonts
\usepackage[utf8]{inputenc}
\usepackage{url}   % simple URL typesetting
\usepackage{booktabs} % professional-quality tables
\usepackage{amsfonts} % blackboard math symbols
\usepackage{nicefrac} % compact symbols for 1/2, etc.
\usepackage{microtype}% microtypography
\usepackage{lipsum}
\usepackage{graphicx}
\graphicspath{ {./figures/} }

\usepackage{siunitx}
\usepackage{multirow}
\usepackage[table]{xcolor}
\usepackage{listings} % Package to display code
\usepackage{xcolor} % Package to add color to code
\usepackage{amsmath}
\usepackage{amssymb}
\usepackage{amsthm}
\usepackage{mathtools}
\usepackage{tikz}
\usepackage{algorithm}%
\usepackage[normalem]{ulem}
\usepackage[noend]{algpseudocode}
\usepackage{filecontents}
\usepackage{xr-hyper}
\usepackage{hyperref} % hyperlinks
\usepackage{cleveref}
\usepackage{nomencl}
\usepackage{etoolbox}
\usepackage{soul}
\usepackage[framemethod=tikz]{mdframed}
\usepackage{verbatim}

\makenomenclature

\makeatletter
\newcommand*{\addFileDependency}[1]{% argument=file name and extension
  \typeout{(#1)}
  \@addtofilelist{#1}
  \IfFileExists{#1}{}{\ypeout{No file #1.}}
}
\makeatother

\makeatletter
\newcommand*{\addAuxFileDependency}[1]{% argument=file name and extension
  \makeatletter\@input{x#1.tex}\makeatother
}
\makeatother

\newcommand*{\myexternaldocument}[1]{%
    \externaldocument[#1:]{#1}%
    \addAuxFileDependency{#1}%
    \addFileDependency{#1.tex}%
    \addFileDependency{#1.aux}%
}

\small

\newcommand{\parsm}{\boldsymbol{\omega_{\mathrm{s}}}}
\newcommand{\pvec}{\mathbf{p}}

\newcommand{\parlumped}{\boldsymbol{\omega}_t}
\newcommand{\opfunbc}{\phi}

\newcommand{\jt}{j_{\mathrm{t}}}

\newcommand{\ld}{{\ell}_{\mathrm{\xvec}}}
\newcommand{\lt}{{\ell}_{\mathrm{t}}}
\newcommand{\lm}{{\ell}_{\mathrm{\dxvec}}}

\newcommand{\sR}{\mathrm{R}}
\newcommand{\sN}{\mathrm{N}}
\newcommand{\kinn}{KINN}

\newcommand{\opnc}{\mathbf{\mathrm{C}_{N}}}

\newcommand {\ic}{IC}
\newcommand {\SA}{Supplementary Material}

\newcommand{\ls}{{\boldsymbol{\ast}}}

\newcommand{\erresp}{\boldsymbol{\varepsilon}}
\newcommand{\cov}{\boldsymbol{\Sigma}}
\newcommand{\icov}{\boldsymbol{\Omega}}
\newcommand{\xvec}{\mathbf{x}}
\newcommand{\dxvec}{\mathbf{\dot{x}}}
\newcommand{\yvec}{\mathbf{y}}
\newcommand{\zvec}{\mathbf{z}}
\newcommand{\cvec}{\mathbf{c}}

\newcommand{\dzvec}{\mathbf{\dot{z}}}
\newcommand{\fmap}{\psi}

\numberwithin{equation}{subsection}

\algrenewcommand{\algorithmicrequire}{\textbf{Input:}}
\algrenewcommand{\algorithmicensure}{\textbf{Output:}}

\renewcommand\nomgroup[1]{%
  \item[\bfseries
  \ifstrequal{#1}{A}{Ordinary Differential Equation}{%
  \ifstrequal{#1}{B}{Surrogate Models}{%
  \ifstrequal{#1}{C}{Residuals}{%
  \ifstrequal{#1}{D}{Parameters}{%
  \ifstrequal{#1}{E}{Dimensionality Reduction}{%
  \ifstrequal{#1}{F}{Abbreviations}{}}}}}}%
]}

\crefname{equation}{Eq.}{Eqs.}
\Crefname{equation}{Equation}{Equations}
\crefname{table}{Table}{Tables}
\Crefname{table}{Table}{Tables}
\crefname{figure}{Fig.}{Figs.}
\Crefname{figure}{Figure}{Figures}
\title{Maximum-likelihood Estimators in Physics-Informed Neural Networks for High-dimensional Inverse Problems}

\author{
  Gabriel S. Gusmão\textsuperscript{a} and Andrew J. ~Medford\textsuperscript{a}\\\\
  \textsuperscript{a}School of Chemical \& Biomolecular Engineering\\
  Georgia Institute of Technology\\
  Atlanta, GA 30332\\
  \texttt{\{gusmaogabriels, ajm\}@gatech.edu}\\\\
}

\myexternaldocument{supp}

\begin{document}

%% %TC: tags to ignore certain parts of the text.
%TC:ignore
% \section*{Word Counts}
% This section is \textit{not} included in the word count.
% \quickwordcount{main}
% \quickcharcount{main}
% \detailtexcount{main}

\nomenclature[A, 01]{$\mathbf{M}$}{Stoichiometry matrix}
\nomenclature[A, 02]{$x$}{State variable}
\nomenclature[A, 03]{$\xvec$}{State variable array}
\nomenclature[A, 04]{$\dxvec$}{State variable time derivatives}
\nomenclature[A, 05]{$\cvec$}{Concentrations vector}
\nomenclature[A, 06]{$\mathbf{r}$}{Rate of reaction vector}
\nomenclature[A, 07]{$\cov$}{Covariance matrix}
\nomenclature[A, 09]{$\icov$}{Precision matrix}
\nomenclature[A, 10]{$\mathbf{\tilde{y}}$}{Unscaled measurements}
\nomenclature[A, 11]{$\mathbf{\tilde{x}}$}{Scaled measurements}
\nomenclature[A, 12]{$\boldsymbol{\gamma}$}{Calibration factor}
\nomenclature[B, 01]{$\opfunbc$}{Activation function}
\nomenclature[B, 02]{$\opnc$}{Normalization-constraint operator}
\nomenclature[B, 03]{$\ell$}{Objective function errors: $\lt,\;\ld,\;\lm$ data and model likelihoods, respectively}
\nomenclature[C, 01]{$\erresp$}{Residuals array}
\nomenclature[C, 02]{$\erresp_i$}{Residuals array at time point $t_i$}
\nomenclature[C, 03]{$\hat{\erresp}$}{Residuals random variable}
\nomenclature[C, 04]{$\delta$}{Sensitivity through total derivative}
\nomenclature[D, 01]{$\pvec$}{Kinetic model parameters}
\nomenclature[D, 02]{$\parsm$}{Surrogate model parameters}
\nomenclature[E, 01]{$\mathbf{U}^\sR$}{Orthogonal basis for the range of $\mathbf{M}$, i.e., $\operatorname{row}[\mathbf{M}]$}
\nomenclature[E, 02]{$\mathbf{U}^\sN$}{Orthogonal basis for the nullspace of $\mathbf{M}$, i.e., $\operatorname{null}[\mathbf{M}]$}
\nomenclature[E, 03]{$\mathbf{V}^\sR$}{Orthogonal basis for the range of $\mathbf{M}^T$, i.e., $\operatorname{row}[\mathbf{M}^T]$}
\nomenclature[E, 04]{$\mathbf{V}^\sN$}{Orthogonal basis for the nullspace of $\mathbf{M}^T$, i.e., $\operatorname{null}[\mathbf{M}^T]$}
\nomenclature[E, 05]{$\mathbf{S}$}{Diagonal matrix with singular values of $\mathbf{M}$}
\nomenclature[F, 01]{SVD}{Singular-value Decomposition}

\maketitle

%%%%%%%%%%%%%%%%%%%%% ABSTRACT %%%%%%%%%%%%%%%%%%%%%%%%%% 
\begin{abstract}
Physics-informed neural networks (PINNs) have proven a suitable mathematical scaffold for solving inverse ordinary (ODE) and partial differential equations (PDE). Typical inverse PINNs are formulated as soft-constrained multi-objective optimization problems with several hyperparameters. In this work, we demonstrate that inverse PINNs can be framed in terms of maximum-likelihood estimators (MLE) to allow explicit error propagation from interpolation to the physical model space through Taylor expansion, without the need of hyperparameter tuning. We explore its application to high-dimensional coupled ODEs constrained by differential algebraic equations that are common in transient chemical and biological kinetics. Furthermore, we show that singular-value decomposition (SVD) of the ODE coupling matrices (reaction stoichiometry matrix) provides reduced uncorrelated subspaces in which PINNs solutions can be represented and over which residuals can be projected. Finally, SVD bases serve as preconditioners for the inversion of covariance matrices in this hyperparameter-free robust application of MLE to ``kinetics-informed neural networks''.

\end{abstract}
%TC:endignore
\section{Main}
\par

Physics-informed~\cite{Raissi2019a} (PINNs), inspired or constrained~\cite{Liu2019Multi-fidelityModeling} neural-networks comprise an interdisciplinary emerging area connecting machine learning to several fields of science, engineering and mathematics. It has shown that neural networks (NNs) provide a flexible structure for the solution of problems arising from fully- or partially-known differential equations, such as in heat-transfer~\cite{Cai2021Physics-informedProblems,Mishra2021PhysicsTransfer}, compressible and incompressible, inviscible and non-inviscible flows~\cite{Jagtap2020ConservativeProblems,Cai2021Physics-informedReview}, advection-diffusion equations~\cite{Pang2019FPinns:Networks}, and other problems involving conservation laws and physical constraints. 

The formulation of inverse problems based on PINNs typically entails softly-constrained cost functions with several penalty weights (hyperparameters) that need be tuned for a given set of data of known underlying physics (differential equation)~\cite{Mishra2021PhysicsTransfer,Jagtap2020ConservativeProblems,Lu2021Physics-InformedDesign,Mishra2022EstimatesPDEs,Yu2022Gradient-enhancedProblems,Chen2021SolvingNetworks,Patel2022ThermodynamicallySystems}. {Other strategies such as weighting based on gradient statistics, inverse Dirichlet weighting, and multiple gradient descent based on the Karush-Kuhn-Tucker (KKT) formulation have been applied~\cite{Du2022InverseNetworks}} to satisfy initial values or boundary conditions. However, it still remains to be devised a general strategy for balancing the contribution between data interpolation and physical model satisfaction. Mean-squared errors (MSE) cost functions are analogous to maximizing the error likelihood if errors are normally distributed, uncorrelated, homoscedastic and of equal variance~\cite{FumioHayashi2000Finite-SampleOLS}. Consequently, the weights of MSE cost-function components in inverse PINNs problems consist of variance-weighted maximum-likelihood estimators (MLE) around uncorrelated homoscedastic normally distributed error terms. The need for supervised screening of such weighting hyperparameters adds a cumbersome layer of complexity to the inverse problem solution, and prevents direct propagation of error between the loss function and the fitted parameters. 

% justification of investigation of PINNs application to high-dimensional inverse problems
Although the applications of PINNs have spanned problems in time and/or spatial (ODEs and PDEs) domains (input spaces) ~\cite{Gao2022Physics-informedProblems,Meng2020,Yu2022Gradient-enhancedProblems,Yang2021B-PINNs:Data,Zubov2021NeuralPDE:Approximations}, few works have assessed inverse problems of high-dimensionality in the state space (targets)~\cite{Ngo2021SolutionNetworks,Yazdaniid2020SystemsDynamics}, which is an intrinsic feature of large interacting systems, such as chemical reactions and biological systems. The few exceptions are the forward solution of the Allen-Cahn, Hamilton–Jacobi–Bellman and Black-Scholes equations~\cite{Han2018SolvingLearningc} in partial stochastic differential equations. Chemical reaction networks may include thousands of chemical species and reactions~\cite{Thybaut2013} in often stiff systems of coupled nonlinear differential equations. Studies involving high-dimensional inverse chemical kinetics problems using standard numerical methods have also been limited~\cite{Berger2008}. They are computationally expensive, requiring successive computation of ODE integrals to estimate gradients through adjoint or sensitivity methods~\cite{Rubert-Nason2014,Rangarajan2017,Marin2021KineticsScaleb}. PINNs might also serve as an auxiliary integration and derivative estimation method in conjunction with mechanism-discovery strategies based on symbolic regression. These include methods that generate many integral-form model candidates, e.g. ALAMO~\cite{Wilson2017TheLearning}, require derivative estimates, e.g. SINDy~\cite{Brunton2016DiscoveringSystems}, or successive numerical integration for validation~\cite{Servia2023TheFrameworks}. Among the few studies involving NN-based approaches to chemical kinetics, De Florio et. al. used the ``extreme theory of functional connection'' to derive upper-bounds for generalization error in stiff forward PINNs. They tested their approach against the \textit{POLLU} problem~\cite{Verwer2006GaussSeidelKinetics}, consisting of a stiff ODE encompassing $20$ chemical species in homogeneous phase. Wu et. al.~\cite{Wu2022PolyODENet:Data} explored the use of lasso-regularized neural ordinary differential equations (NODE) to enforce sparsity in inferring the mass-action rate law coupling matrix for homogeneous chemical systems involving six chemical species. However, studies involving PINNs for inverse complex reaction systems, especially in heterogeneous catalysis, are still scarce \cite{Gusmao2022Kinetics-informedNetworksb}. Such reacting systems are ubiquitous in the chemical industry and the ability to investigate the detailed mechanisms and elementary processes that underlie chemical reactions would allow the rigorous modeling, control and design of chemical reactors, and the optimization of industrial processes.

% introduction to KINNs, MLE and SVD preconditioning
Kinetics-informed neural networks, KINNs, have been proposed by the authors and co-workers as a particular application of PINNs to high-dimensional heterogeneous chemical kinetics, which structurally tackles the issue of differential-algebraic equation (DAE) constraints and estimation of derivatives across the NN structure~\cite{Gusmao2022Kinetics-informedNetworksb}. However, one of the main weakness of the original inverse KINNs formulation is that, as inverse PINNs in general, it consists of a multi-objective optimization problem. It required a priori estimation of a hyperparameter that conveyed the ratio between variances of state values and their derivatives. The inverse solution was shown to be a function of the chosen hyperparameter, leading to a Pareto frontiers of efficient solutions. A more effective approach is necessary to rigorously model the probability density function of residuals. The MLE formulation is advantageous since it may be applied to complex error distributions, e.g., full covariance with correlated error terms for Gaussian error~\cite{Biegler1986NonlinearComparison}, and its structure  allows the propagation of variance between residuals. However, MLEs constructed around multivariate Gaussian distributions require the inversion of error covariance matrices, and correlated errors lead to ill-conditioning with respect to inversion. Better conditioning can be attained by finding a suitable preconditioner for the coupled differential equations~\cite{Wathen2015Preconditioning}. The ODEs that describe the rates of change of species in a chemical system are dictated by material balances (conservation laws), which are conveyed as reaction stoichiometries~\cite{Dumesic1993,Constales2014TheStructure,Redekop2014ElucidatingData}. For large chemical systems, these linear relationships between changes in concentrations of chemical species lead to inherently correlated errors in states and their derivatives, which results in ill-conditioned covariance matrices.

% summary of this work
In this work, we show that the Taylor expansion of residuals in Gaussian MLEs allows direct error propagation between interpolation and physics models for high-dimensional ODE inverse problems, thereby removing the need for hyperparameter tuning in inverse PINNs, especially for large chemical systems (\kinn{s}). Furthermore, we resort to the singular-value decomposition (SVD) as a framework to generate orthonormal bases for the range and nullspace of the ODE coupling matrix (reaction stoichiometry matrix). The range basis serves as preconditioner since the projection of KINNs residuals on it removes correlated terms from the error covariance matrix, which mitigate the issue of ill-conditioning with respect to inversion, allowing the estimation of precision matrices. In this robust KINNs formulation (rKINNs) we incorporate the three types of learning biases, as defined by Karniadakis et. al.~\cite{Karniadakis2021Physics-informedLearning}: (i) ``observational bias'' in the sampling of model error at points where observed data is not available, (ii) ``inductive bias'' in the conservation law defined by the stoichiometry matrix, the topological incorporation of the normalization operator that enforces DAE constraints, as well as in the representation of the ODEs in the preconditioning space obtained from SVD of the stoichiometry matrix, and (iii) ``learning bias'' in assuming model and interpolation residuals are normally distributed and constructing the optimization problem in terms of MLEs.
\section{Results}

\subsection{Theoretical Overview}

\subsubsection{Mean-field chemical kinetics}
\label{subsec:mf_mkm_fmwrk}
In its conceptualization, KINNs was designed around mean-field kinetic models with power-law kinetics with Arrhenius temperature, $\theta$, dependence for a set of chemical species whose concentrations (states) are given by an array $\mathbf{c}\in\mathbb{R}^{n}_+$. A mean-field microkinetic model (MKM) is defined as in \eqref{eq:kin_ode}:
\begin{equation}
    \mathbf{\dot{c}}=\frac{d}{dt}\mathbf{c}=f(\mathbf{c},\pvec(\theta))= \mathbf{M}\,\mathbf{r}(\mathbf{c},\pvec(\theta))= \mathbf{M}\left(\mathbf{k}(\pvec(\theta))\circ \fmap(\mathbf{c})\right)\label{eq:kin_ode}
\end{equation}
Where $\mathbf{M}\in\mathbb{Z}^{n\times m}$ is the stoichiometry matrix (ODE coupling matrix), which can include reactions between chemical species possibly in different phases (e.g. surface and gas/liquid-phases), $~{\fmap(\cdot):\mathbb{R}^n_+\to\mathbb{R}^m_+}$ represents the power-law kinetics mapping, and $~{\mathbf{k}:=\{\mathbf{k}=\exp(-\pvec(\theta))~\in~\mathbb{R}^m_+,\,\theta\in\mathbb{R}_{+},\, \pvec(\theta)\in\mathbb{R}^m\}}$ denotes Arrhenius-like temperature dependent rate constant with $\pvec$ as a vector-valued function that includes entropic and enthalpic contributions to a transition state. 
% Equation \eqref{eq:kin_ode} is analogous to a closed pseudo-homogeneous ideal batch reactor (uniform properties), which is the core representation of the dynamics of a chemical transformation in a reactive system, and it will be used as base case for the further analyses in this study. The ideal batch model can, however, be converted to open systems, such as continuous-flow stirred tank (CFSTR) reactors or steady-state plug-flow (PFR) or packed-bed reactors (PBR) with minor modifications. For the CFSTR case, it would require including an input stream and assuming the output stream to have the same concentration of the known reactor volume, whereas for steady-state PFRs or PBRs, the time derivative is replaced by a spatial (or per-catalyst mass) derivative. Additional modifications would allow the equation to be adapted to other systems where heat and mass transport resistance involved; however, the core complexity in chemical transfomations can be generally expressed by \cref{eq:kin_ode}.  
%
\subsubsection{Inverse KINNs residuals}
``Surrogate approximator'' (SA) were introduced in KINNs~\cite{Gusmao2022Kinetics-informedNetworksb} as a composite set of NNs that serve as basis functions with $\parsm$ weights for the solution of chemical kinetics ODEs, \cref{eq:kin_ode}, with $\mathbf{c}(t)\approx\xvec(t,\parsm)$, with the ability to structurally represent DAE constraints through an operator $\opnc$ (\cref{subsec:nn_representation}). Such an approach builds upon the standard PINNs concept~\cite{Meade1994c,Meade1994b,Lagaris1998a}, which is a direct result of the universal approximation theorem~\cite{Cybenko1989,Hornik1990,Hornik1991,Leshno1993}. The inverse problem is defined in terms of the multi-objective optimization function involving interpolation and physics-model residuals, $\erresp_{\xvec_i}$ and $\erresp_{\dxvec_i}\in\mathbb{R}^n$ for $i=1,2,...,d$ datapoints, respectively. The interpolation error, \cref{eq:res_interp}, is defined by the difference between measured states values, $\tilde{\xvec}_i\in\mathbb{R}^n$, and estimated state values, $\xvec(t_i,\parsm)\in\mathbb{R}^n$, for every $t_i$:
\begin{align}
    {\erresp}_{\xvec_i}(\parsm) &= \xvec(t_i,\parsm)-\tilde{\xvec}_i\label{eq:res_interp}
\end{align}
such that if $\tilde{\erresp}_{\xvec_i}(\parsm) = 0$ then the SA goes exactly through data point $i$. The physics-model residual, $\erresp_{\dxvec_i}$, relates the SA derivative and the output of differential equation (e.g. the kinetic model) in \cref{eq:res_model}.
\begin{equation}
\begin{aligned}
\erresp_{\dxvec_i} &=\dxvec(t_i,\parsm)- f(\xvec_i, \pvec) \\ 
    &= \dxvec(t_i,\parsm)- \mathbf{M}\left(\mathbf{k}(\pvec(\theta))\circ \fmap(\xvec(t_i,\parsm))\right)\label{eq:res_model}
\end{aligned}
\end{equation}
such that if $ \erresp_{\dxvec_i} = 0$ then the differential equation in \cref{eq:kin_ode} is exactly satisfied at time $t_i$. 

In the original regularized \kinn{s} loss function, a hyperparameter $\alpha$ weights the interpolation in relation to the model-fitting MSE, \cref{eq:regularized}.
\begin{equation}
    \begin{aligned}    \jt(\parsm, \pvec)\operatorname{MSE}({\erresp_{\dot{\xvec}}})+\alpha \operatorname{MSE}({\erresp_{{\xvec}}})=\frac{1}{d}\sum_i^d\erresp_{\dot{\xvec}_i}^T\erresp_{\dxvec_i}+\alpha\frac{1}{d}\sum_i^d\erresp_{\xvec_i}^T\erresp_{\xvec_i}
    \end{aligned}
    \label{eq:regularized}
\end{equation}
The loss function, $\jt$ is then minimized with respect to $\parsm$ and $\pvec$ for a fixed $\alpha$. The error in kinetic model parameters, $\erresp_{\pvec}$, is defined as:
\begin{equation}
    \erresp_{\pvec}^* = \pvec - \pvec^*
    \label{eq:err_pvec}
\end{equation}
where $\pvec$ represents the parameters of the physical model at any point in the optimization and $\pvec^*$ represents the true model parameters. Unlike  ${\erresp}_{\xvec_i}$ and $\erresp_{\dxvec_i}$, $\erresp_{\pvec}^*$ is not dependent on time, and cannot be evaluated explicitly from kinetic data since $\pvec^*$ is not known in general. Rather, it is implicitly assumed that minimizing the \kinn{s} loss function also minimizes $\erresp_{\pvec}^*$. 
\subsection{Derivation of maximum-likelihood estimator}

\subsubsection{\textbf{Error probability density function}}To circumvent the need for a regularization hyperparameter ($\alpha$) and enable rigorous error propagation, the original \kinn{s} problem, \cref{eq:regularized}, and most or all related PINN problems, can be reformulated in terms of maximum likelihood estimators, MLEs. The key idea of MLE is to assume an underlying probability distribution function, $\pi(\varepsilon)$, for the error terms in the regression problem and to maximize the likelihood of observing the errors given the data. Let $\erresp_{\xvec}$, $\erresp_{{\dxvec}}$, $\erresp_{\pvec}$ denote the errors related to the interpolation, physical model and its parameters, respectively. Assuming errors are independent between timepoints $i$, and that $\erresp_\xvec$ is normally distributed with covariance $\cov_\xvec$, the associated probability distribution function conditioned to the observed data $\tilde{\xvec}$ and its covariance $\cov_\xvec$ is given in \cref{eq:pdf}.
\begin{align}
    \pi(\cap_{i=1}^n\erresp_{\xvec_i},\erresp_{\dot{\xvec}_i},\erresp_{\pvec}|\cov_{\xvec},\mathbf{\tilde{x}}_i)=\prod_i^n\pi(\erresp_{\xvec_i},\erresp_{\dxvec_i},\erresp_{\pvec}|\cov_{\xvec},\mathbf{\tilde{x}}_i)\label{eq:pdf}
\end{align}
 The main advantage of the MLE formulation to KINNs is that it allows direct connection between errors associated with model-fitting and those related to data interpolation. With the independent variables $\xvec$ and $\pvec$, and assuming $\dxvec(\xvec)$ through $\parsm$, a variation in $\erresp_\dxvec$ can be represented by its total derivative:
\begin{equation}
        % \delta\erresp_\dxvec=\left.\underbrace{\partial_\xvec\dxvec \delta\xvec}_{\mathclap{\text{representation}}}\right.-\left.\underbrace{\left(\partial_\xvec f(\xvec,\pvec){\delta}\xvec+\partial_{\pvec}f(\xvec,\pvec){\delta}{\pvec}\right)}_\mathclap{\text{structural}}\right.+\mathcal{O}(\delta\xvec^2,\delta\pvec^2) \label{eq:linearized}
        \delta\erresp_\dxvec=\underbrace{\partial_\xvec\dxvec \delta\xvec}_{\mathclap{\text{representation}}}-\left.\underbrace{\left(\partial_\xvec f(\xvec,\pvec){\delta}\xvec+\partial_{\pvec}f(\xvec,\pvec){\delta}{\pvec}\right)}_{\text{structural}}\right.+\mathcal{O}(\delta\xvec^2,\delta\pvec^2) \label{eq:linearized}
\end{equation}
which represents the sensitivity of the model residuals, $\delta\erresp_\dxvec$, in a Taylor expansion as a function of variations in model parameter, $\delta\pvec$, and states, $\delta\xvec$. In \cref{eq:linearized}, $\partial_\xvec f(\xvec,{\pvec})$ and $\partial_{\pvec}f(\xvec,{\pvec})$ are the kinetic model gradients with respect to states and physical model parameters, respectively, about $\xvec$ and ${\pvec}$, and they represent the contribution of the physical model to the variations in the residuals of the SA. The partial derivatives between the time derivatives of the SA with respect to its states, $\partial_\xvec\dxvec$, are implicit functions of the NN architecture, i.e., depth, width, connectivity, choice of activation functions and, ultimately, their weights, $\parsm$, as in \cref{eq:ddxdx}. 
\begin{align}
    \partial_\xvec\dxvec&=\frac{\partial\dxvec}{\partial\parsm}\frac{\partial\parsm}{\partial\xvec}=\frac{\partial\dxvec}{\partial\parsm}\left(\frac{\partial\xvec}{\partial\parsm}\right)^{-1}\label{eq:ddxdx}
\end{align}
The universal approximation theorem extends the approximation of functions not only to their values but their gradients~\cite{Pinkus1999ApproximationNetworks,Mai-Duy2003ApproximationNetworks}. Therefore, in the limit of sufficiently complex NNs, the models states, $\xvec$, and the model derivatives, $\dxvec$, will be independent. This is analogous to the case of explicit Galerkin methods that employ bases which enforce orthogonality between states and associated derivatives~\cite{Livermore2009GalerkinPolynomials,Livermore2010Quasi-LExpansions}. However, for smaller NNs there should exist some correlation between states and their derivatives, with the ideal scenario being $\partial_\xvec\dxvec\propto\partial_\xvec f$, where SA behavior mimics that of the physical model making a PINN approximate a NODE.
Nonetheless, in this work we neglect the representation error, $\partial_\xvec\dxvec$, based on the assumption that the feed-forward NN basis is large enough to allow states and their derivatives to be represented independently. With the assumption of normality distribution of residuals, the derivative residuals covariance matrix at a timepoint $i$, $\cov_{\dxvec_i}$ can be expressed as projections of the interpolation covariance onto the gradients of the model at the respective timepoints, as in \cref{subsec:pdf_assumptions}. The joint probability density function (PDF) for the MLE approach is then given by the product of independent distributions: 
\begin{align}
    \pi\left(\cap_{i=1}^n\erresp_{\xvec_i},\erresp_{\dot{\xvec}_i},\erresp_{\pvec}|\cov_{\xvec},\mathbf{\tilde{x}}_i\right)&\approx\;\pi(\cap_{i=1}^n\erresp_{\xvec_i},\erresp_{\dot{\xvec}_i}|\cov_{\xvec},\cov_{\dxvec_i},\mathbf{\tilde{x}}_i,\pvec)\\
    &=\prod_i^n\pi(\erresp_{\dot{\xvec}_i}|\cov_{\dxvec_i})\pi(\erresp_{\xvec_i}|\cov_{\xvec})\label{eq:likhd}
\end{align}
In \cref{eq:likhd}, we imply that model derivative and interpolation covariances are assumed constant for a given sample of evaluated model and model derivative errors. In \cref{subsec:mle}, we show how this assumption about the statistical model can  be used within a maximum-likelihood estimator framework.

\subsubsection{\textbf{Maximum likelihood estimation}}\label{subsec:mle} Maximizing likelihood estimators are equivalent to minimizing the negative log-likelihood of error PDFs~\cite{Hald1999OnSquares,Myung2003TutorialEstimation}. In MLE form, \cref{eq:likhd} represents a summation, as in \cref{eq:loglikhd}, where $\pi$ are assumed to be Gaussian PDFs  and $\icov=\cov^{-1}$ is the precision matrix, which is held constant within parameter optimization and, therefore, does not depend on the suppressed variables $\parsm$ and $\pvec$. For brevity, we lump all parameters being optimized into a single variable, $\parlumped = [\parsm, \pvec]$, yielding.
\begin{equation}
\begin{aligned}
    \min_{\parlumped} \quad &\lt=\jt=\frac{1}{d}\sum_i^d\left(\erresp_{\dot{\xvec}_i}^T\icov_{\dxvec_i}\erresp_{\dxvec_i}+\erresp_{\xvec_i}^T\icov_{\xvec}\erresp_{\xvec_i}\right) \\
    \textrm{s.t.}\quad&\icov_{\dxvec_i}=\left(\cov^\xvec_{\dxvec_i}+\cov^{\pvec}_{\dxvec_i}\right)^{-1}\\
    \quad&\icov_\xvec\;=\cov_{\xvec}^{-1}\\
    \quad&\pvec\in\mathbb{R}^{\operatorname{dim}(\pvec)},\;\parsm\in\mathbb{R}^{\operatorname{dim}(\parsm)}
    \label{eq:loglikhd}
\end{aligned}    
\end{equation}
\subsection{Range-Nullspace Decomposition}
Coupled ODEs describing rank-deficient dynamical system can be reduced by defining proper bases for states and their derivatives. The algebraic analysis of the stoichiometry coupling matrix for information recovery and data reconstructions has been focus of investigation in chemical and biological systems~\cite{MacDougall1937ThermodynamicRysselberghe,Redekop2011}. Here we generalize the algebraic analysis of the coupling matrix, $\mathbf{M}$ through it singular-value decomposition (SVD), which is determined by \cref{eq:svd}.
\begin{equation}
    \mathbf{M}\triangleq\mathbf{U}\mathbf{S}\mathbf{V}^\dag\label{eq:svd}
\end{equation}
Where $\mathbf{U}\in\mathbb{R}^{n\times n}$, $\mathbf{V}\in\mathbb{R}^{m\times m}$ are orthonormal basis of $\mathbb{R}^n$ and $\mathbb{R}^m$, respectively, and $\mathbf{S}\in\mathbb{R}^{n\times m}$ is a matrix with singular values of $\mathbf{M}$ along its diagonal entries and zeroes for the off-diagonal ones. Therefore, $\mathbf{U}$ is a basis for concentrations and their derivatives over time, which can be broken down into $\mathbf{U}^{\sR}=\{u_{ij}\;|\;s_{jj}>0,\;\forall i\}$, which is a basis for the range of $\mathbf{M}$, and $\mathbf{U}^{\sN}=\{u_{ij}\;|\;s_{jj}=0,\;\forall i\}$, which is a basis for the null-space or kernel of $\mathbf{M}^T$.
\begin{equation}
    \mathbf{U}=\begin{bmatrix}\mathbf{U}^{\sR}&\mathbf{U}^{\sN}\end{bmatrix}\label{eq:de_reac_red}
\end{equation}
Any net non-zero rate of reaction must be in $\mathbf{U}^{\sR}$, while the boundary conditions and inner fluxes within the chemical system lie in the null-space $\mathbf{U}^{\sN}$, allowing the concentration vector to be decomposed into a range and nullspace component:
\begin{align}
    \cvec_{(t)}&=\cvec^{\sR}_{(t)}+\cvec^{\sN}\label{eq:conc_rn}\\
    \cvec^{\sR}_{(t)}&=\mathbf{U}^{\sR}(\mathbf{U}^{\sR})^T\cvec_{(t)}=\mathbf{U}^{\sR}\mathbf{z}^{\sR}_{(t)}\label{eq:crange}\\
    \cvec^{\sN}&=\mathbf{U}^{\sN}(\mathbf{U}^{\sN})^T\cvec_{(t)}=\mathbf{U}^{\sN}\mathbf{z}^{\sN}\label{eq:cnull}
\end{align}
where $\mathbf{z}^{\sN}_{(t)}$ and $\mathbf{z}^{\sR}_{(t)}$ are the subspace projections of $\cvec$ onto the null-space and range, respectively. As for concentration derivatives, they exclusively reside in the range of $\mathbf{M}$, and can be decomposed using the projections as follows.
\begin{align}
    \mathbf{\dot{c}}_{(t)}&=\mathbf{\dot{c}}^{\sR}_{(t)}=\mathbf{U}^{\sR}(\mathbf{U}^{\sR})^T\mathbf{\dot{c}}_{(t)}=\mathbf{U}^{\sR}\mathbf{\dot{z}}^{\sR}_{(t)}\label{eq:dcrange}\\
\mathbf{0}&=(\mathbf{U}^{\sN})^T\mathbf{\dot{c}}_{(t)}=\mathbf{\dot{z}}^{\sN}\label{eq:dcnull}
\end{align}
The non-zero singular values from SVD are associated with the subspace over which all the model derivative variance is confined. In \cref{subsec:rkin_mle}, we show that the SA can be built such that the time dependence is only contained in the span of $\mathbf{U}^\sR$, with a time-independent shift in $\mathbf{U}^\sN$. Hence, the projection of \cref{eq:res_interp} onto $\mathbf{U}^\sN$ will lead to a time-independent error $\erresp_\zvec^\sN\in\mathbf{U}^\sN$ between the SA estimates and the observed data. Assuming that $\mathbf{z}^\sN$ can be estimated from the observed data, the nullspace solution can be set a priori as $\mathbf{z}^\sN=\langle\left(\mathbf{U}^\sN\right)^T\tilde{\mathbf{x}}\rangle$. In \cref{eq:loglikhd}, it is assumed that $\icov_\xvec$ can be defined as $\cov^{-1}_\xvec$; however, when $\mathbf{M}$ has zero singular values, $\cov_\xvec$ is rank-deficient and, hence, ill-conditioned with respect to inversion. Projecting $\cov_\xvec$ onto $\mathbf{U}^\sR$ provides a route to preserve the total variance and allows the inversion of the projected matrix, i.e, if $\left(\mathbf{U}^\sR\right)^T\cov_\xvec\mathbf{U}^\sR=\cov_\zvec$ is full-rank, $\icov_\zvec=\left(\cov_\zvec\right)^{-1}$ can be computed. The same rationale holds for $\cov_\dxvec$, since derivative residuals lie only in $\mathbf{U}^\sR$.
\subsection{Case Studies and Discussions}
\label{sec:results}
% In the following section, we explore the outcomes of the MLE formulation of KINNs coupled with propagation of errors from the interpolation to physics model space, and compare it to the previous naïve MSE multi-objective optimization under hyperparameter screening~\cite{Gusmao2022Kinetics-informedNetworksb}. Furthermore, we employ the closed-form solution for catalyst state reconstruction, \cref{supp:sec:data_recnstrctn}, which is analyzed for the scenario of ``deep'' reaction networks with reactions between latent species. %, where there exists at least one reaction between radicals on the catalyst surface.
%
\subsubsection{Representative `Deep' Microkinetic Model}\label{sec:data_generation}
Here, we utilize the most complex of the anecdotal test cases from the prior KINNs study~\cite{Gusmao2022Kinetics-informedNetworksb}, which includes latent species and reactions between them. The elementary steps are given in \cref{eq:rxns}, and the corresponding matrix representation is shown in \SA, \ref{supp:sec:structure}. 

{Heterogeneous Reaction Network}
\begin{equation}
\begin{aligned}
    A+*&\underset{k_{\text{-}1}^d}{\stackrel{k_1^d}{\rightleftharpoons}} A*\\
    B+*&\underset{k_{\text{-}2}^d}{\stackrel{k_2^d}{\rightleftharpoons}} B*\\
    C+*&\underset{k_{\text{-}3}^d}{\stackrel{k_3^d}{\rightleftharpoons}} C*\\
    A*+*&\underset{k_{\text{-}3}^c}{\stackrel{k_3^c}{\rightleftharpoons}} 2D*\\
    B*+*&\underset{k_{\text{-}4}^c}{\stackrel{k_4^c}{\rightleftharpoons}} 2E*\\
    D*\:+\:E*&\underset{k_{\text{-}1}^s}{\stackrel{k_{1}^s}{\rightleftharpoons}} F*\:+\:*\\
    F*\:+\:E*&\underset{k_{\text{-}5}^c}{\stackrel{k_5^c}{\rightleftharpoons}} C*\:+\:*\\
\end{aligned}\label{eq:rxns}
\end{equation}\\

\begin{figure}[hbt!]
    \centering
    \includegraphics[keepaspectratio=true,scale=1.,clip=True]{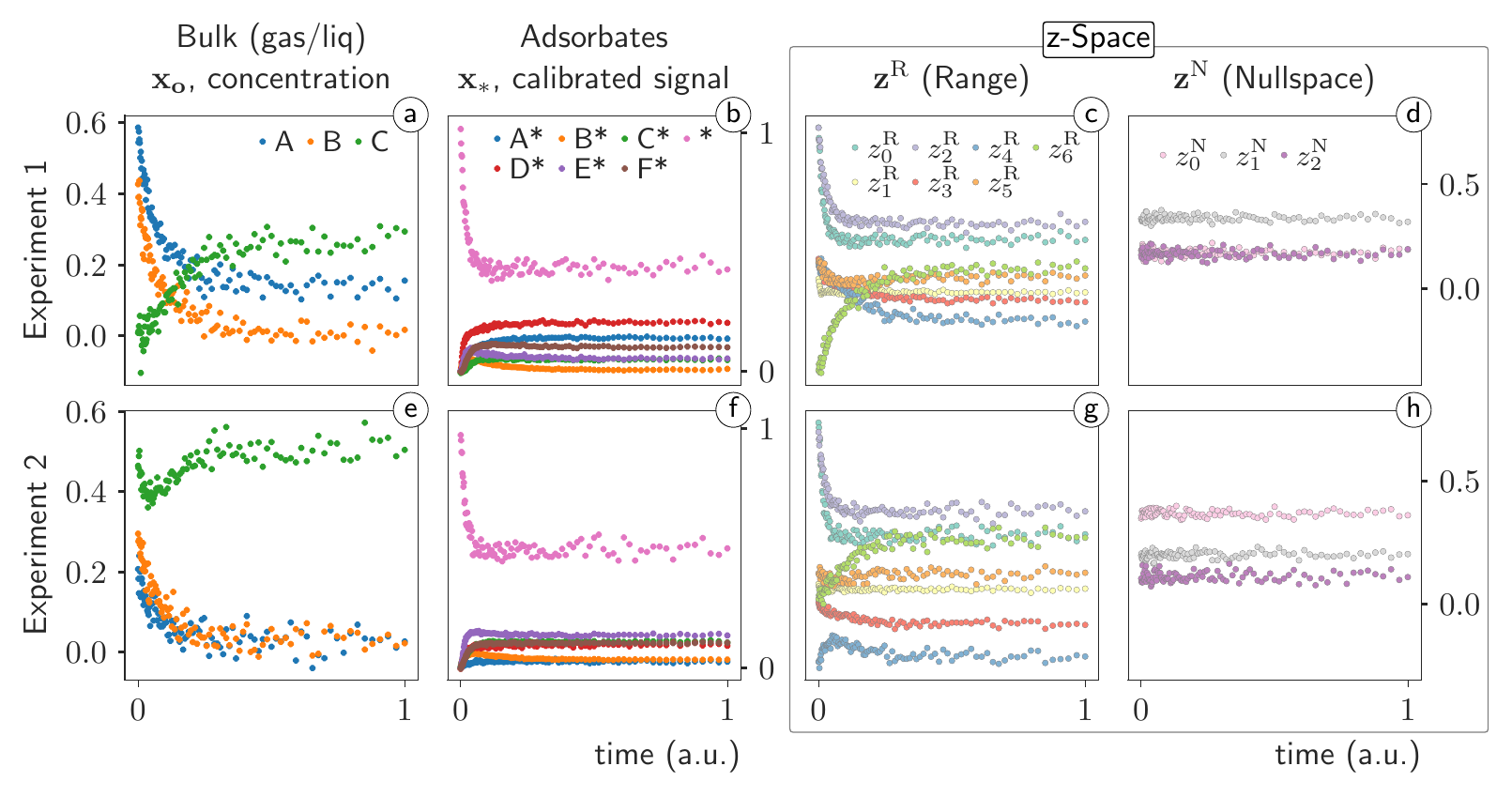}
    \caption{Bulk phase (a and e) and calibrated surface coverage fraction (b and f); $\zvec$-Space: range, $\zvec^\sR$ (c and g), and nullspace, $\zvec^\sN$ (d and h), projections, where all variance is contained in the range projection and time invariance in the nullspace, for \ic 1, $\{x_A,x_B,x_C,x_*\}_{t_0}=\{0.6,0.4,0.0,1.0\}$, and \ic 2, $\{x_A,x_B,x_C,x_*\}_{t_0}=\{0.2,0.3,0.5,1.0\}$.}\label{fig:zspace}
\end{figure}

An overview of the generated synthetic data is shown in \cref{fig:zspace}, where each initial condition is referred to as an ``experiment'', and data is separated by gas-phase concentrations ($\xvec_\mathbf{o}$) that can be directly measured and surface concentrations ($\xvec_\mathbf\ls$) that can only be measured indirectly, which is a common challenge in the dynamics of heterogeneous catalytic systems, i.e. the existence of ``latent'' states arising from adsorbed surface species. The calibrated signals for surface concentrations were obtained from the raw synthetic data by using the closed-form solution for surface reconstruction defined in \cref{supp:eq:null_min}, \cref{supp:sec:data_recnstrctn}, with an eigenvalue cutoff of $5\times10^{-3}$ to find the calibration coefficients $\mathbf{\gamma}$, which are shown in the parity plot in \cref{supp:fig:calibration}. The coverages are normalized by construction, according to \cref{supp:eq:opt_norm}, and satisfy the nullspace time invariance, which expressed by the differential equation in $z$-space, \cref{fig:zspace}. Therefore, for the synthetic experiments under study, the bulk- and surface-phases dynamics can be reconstructed directly rather than learned through NN training.
~~
\subsubsection{Application and Comparison of naïve \kinn{s} and robust KINNs}\label{sec:inverse_rkinns}
% overview and NN structures
The range-nullspace decomposition in \cref{fig:zspace} leads to the dimensionality reduction in the inverse problem, from a total of initially nine degrees of freedom, i.e. three bulk-phase species, and six surface species (adsorbates), while the free-sites balance ``$*$'' degree-of-freedom is accounted for by the normalization operator $\opnc$, only the seven range components are taken into consideration in the optimization. Therefore, the \kinn{s} output layer included three linear components for bulk phases and seven DAE-constrained components with underlying NN with six outputs for surface species. r\kinn{s}' SA only maps the seven components of the range space, while the three nullspace components are a priori estimated. 

\begin{figure}[ht!]
    \centering
    \includegraphics[keepaspectratio=true,scale=1.,clip=True]{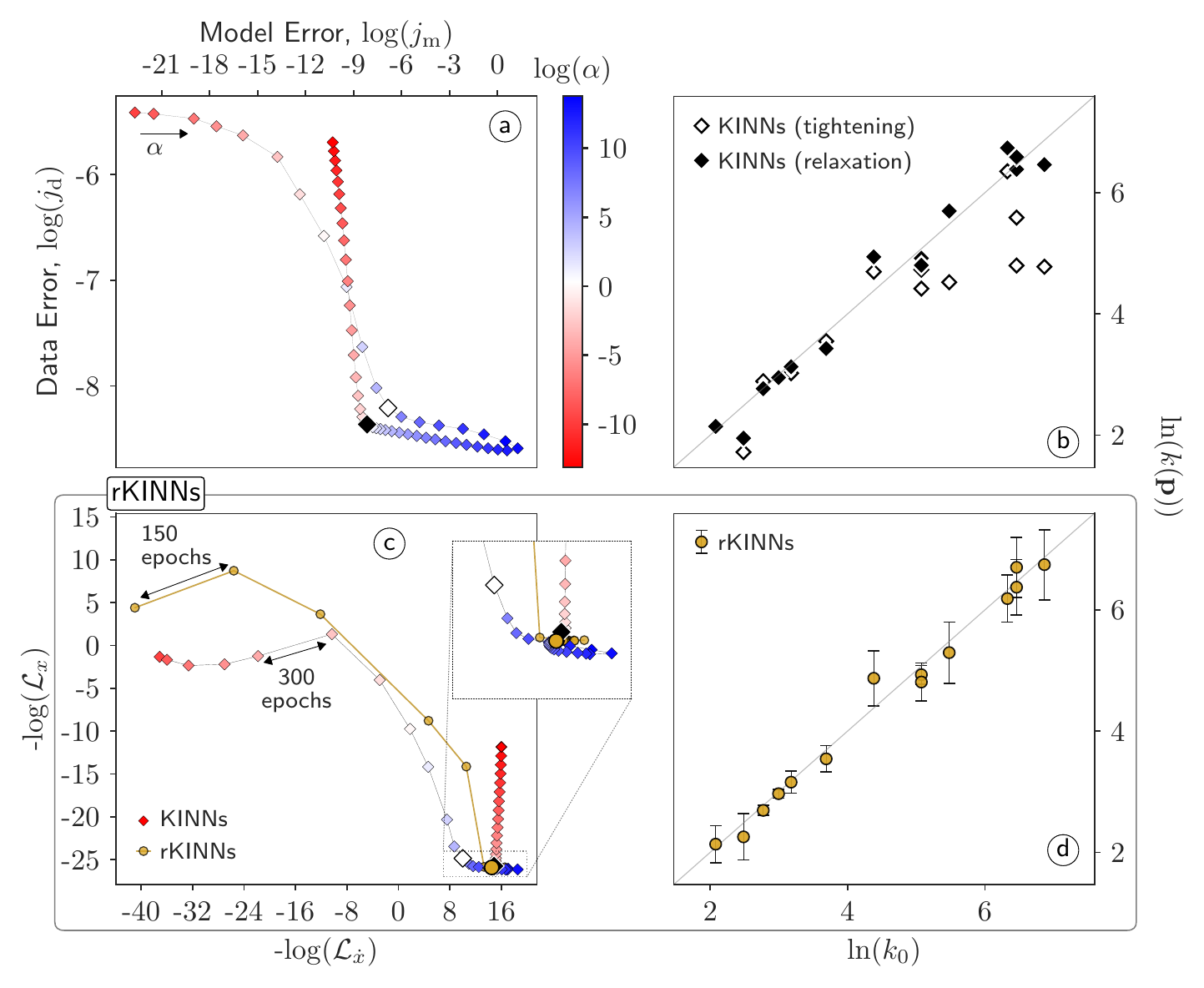}%wid\thetah=1.\linewid\thetah
    \caption{Inverse {\kinn}s vs r{\kinn}s~- (a) sensitivity analysis of naïve {\kinn}s regularization parameter $\alpha$ with Pareto set estimate as a function of interpolation and model MSEs; highlighted tightening (increasing $\alpha$, open diamond) and relaxation (decreasing $\alpha$ closed diamond) inflection points; (b) naïve \kinn{s} model parameters parity plot at inflection points in highlighted in (a); (c) naïve \kinn{s} residuals represented in terms of negative log-likelihoods ($\alpha$-colored circles) and r\kinn{s} (open circles) log-likelihoods reconstructed in terms of the final stable point (black-filled circle in the inset of (c)); (d) r\kinn{s} model parameters at the stable point (error bars defined as $2$ standard deviations based on the Fischer information estimated through Hessian analysis with algorithmic differentiation). One epoch is defined as $100$ parameter-update iterations.}\label{fig:inv_pareto_mle_short}
\end{figure}

% naive approach results and review
In the naïve approach, \kinn{s}, the hyperparameter $\alpha$ controls the expected ratio between interpolation (data) and model (physics) variances. From the MLE perspective, this represents a (naïve) assumption that the underlying interpolation and model residuals are normally distributed with covariances $\cov_\xvec=\sigma_\xvec\mathbf{I}$ and $\cov_\dxvec=\sigma_\dxvec\mathbf{I}$, with $\alpha=\sigma_\dxvec/\sigma_\xvec$, such that smaller values of $\alpha$ denote lower variance for the model residual as compared to the interpolation residuals, and the opposite for larger values of $\alpha$. In \cref{fig:inv_pareto_mle_short} (a), the inverse naïve-\kinn{s} are applied to the case-study system. The data- and model-residuals evolution as a function of $\alpha$ define a Pareto-set of efficient solutions. A parity plot for the kinetic model parameters at the inflection points depicted by open (tightening) and closed (relaxation) diamonds in \cref{fig:inv_pareto_mle_short} (a) is shown in (b). Optimal kinetic model parameters are associated with the ``elbows'' or inflection points in the relaxation direction. Increasing $\alpha$ to larger values leads to SA overfitting to data, whereas reducing it indefinitely leads to SA ``forgetting'' the data, which leads to a scenario of producing degenerate physics solutions~\cite{Gusmao2022Kinetics-informedNetworksb}. The process of optimizing $\alpha$ through multi-objective optimization is computationally expensive, since each iteration requires 300 epochs of NN training, and leads to ambiguity in the solution due to hysteresis in the parameters as a function of $\alpha$.

% naive and robust kinns from MLE perspective
\cref{fig:inv_pareto_mle_short} (c) illustrates the main advantage of r\kinn{s} over \kinn{s}. Interpolation ($\ld$) and model ($\lm$) negative log likelihoods were computed from the interpolation residuals projected onto the final (stationary) covariance matrix obtained by r\kinn{s}, and model residuals per time point according to \cref{eq:loglikhd_rn}, respectively. In addition to the faster per-epoch convergence, r\kinn{s} reaches a stable or stationary point (\cref{fig:inv_pareto_mle_short} (e), black-filled circle). Such a behavior is attributed to the stationarity in the SA interpolation and, thus, of the sample covariance, $\hat{\Sigma}_\xvec$, and its propagation through the model gradients, $\hat{\Sigma}_{\dxvec_i}$, over time points $t_i$. Different choices of hidden-layers activation functions sets have led to indistinguishable results for stationary model parameters and confidence intervals for r\kinn{s}, e.g., $\{20_{\times3}\}$ full swish $\{swish_{\times 3}\}$, $\{100_{\times1}\}$ with $\{rbf\}$, thus highlighting the robustness of r\kinn{s} to not only weight the variances between interpolation and physics, but also to converge to similar convex basins irrespective of the choice of activation functions (basis functions) and SA architecture.

% The r\kinn{s} stability at its stationary point can be further assessed based on the MLE-form convexity for given estimate for $\xvec\approx\tilde\xvec$ and model parameter solutions $\pvec$. The eigenvalue analysis of the Hessian matrix determines the convexity of model parameters at a particular point in the domain~\cite{Nocedal2006NumericalOptimization}. For r\kinn{s}, the Hessian matrix can be efficiently evaluated with JAX algorithmic differentiation. Moreover, the MLE construction of r\kinn{s} allows the estimation of the model parameters' asymptotic covariance matrix through the Fisher information matrix, $\cov_\pvec\propto \mathcal{I}(\pvec)^{-1}$, which is connected to the Hessian matrix of the cost function $\mathcal{I}(\pvec)=\nabla_\pvec^2(\lt)$~\cite{Kalmikov2014AEstimation,Zhan2022UncertaintyModels}. The uncertainty in the calibration factors obtained using the closed-form solution \cref{supp:eq:null_min}, \cref{supp:sec:data_recnstrctn}, need also be accounted for and thus propagated to model parameter uncertainty since they affect the scale of the range equations and lead to linear shifts in the nullspace. As a result, $\cov_\pvec$ expresses the conditional model parameters covariance matrix given the uncertainty in the calibration coefficients $\boldsymbol{\gamma}$ (\SA, \ref{supp:sec:conditional_covar}). The $95\%$-confidence intervals ($\pvec\pm2\cov_\pvec$) are expressed in \cref{fig:inv_pareto_mle_short} (d)  for $\cov_\pvec=\operatorname{diag}(\cov_\pvec)^{1/2}$.
%
\begin{figure}[ht!]
    \centering
    \includegraphics[keepaspectratio=true,scale=1.,clip=True]{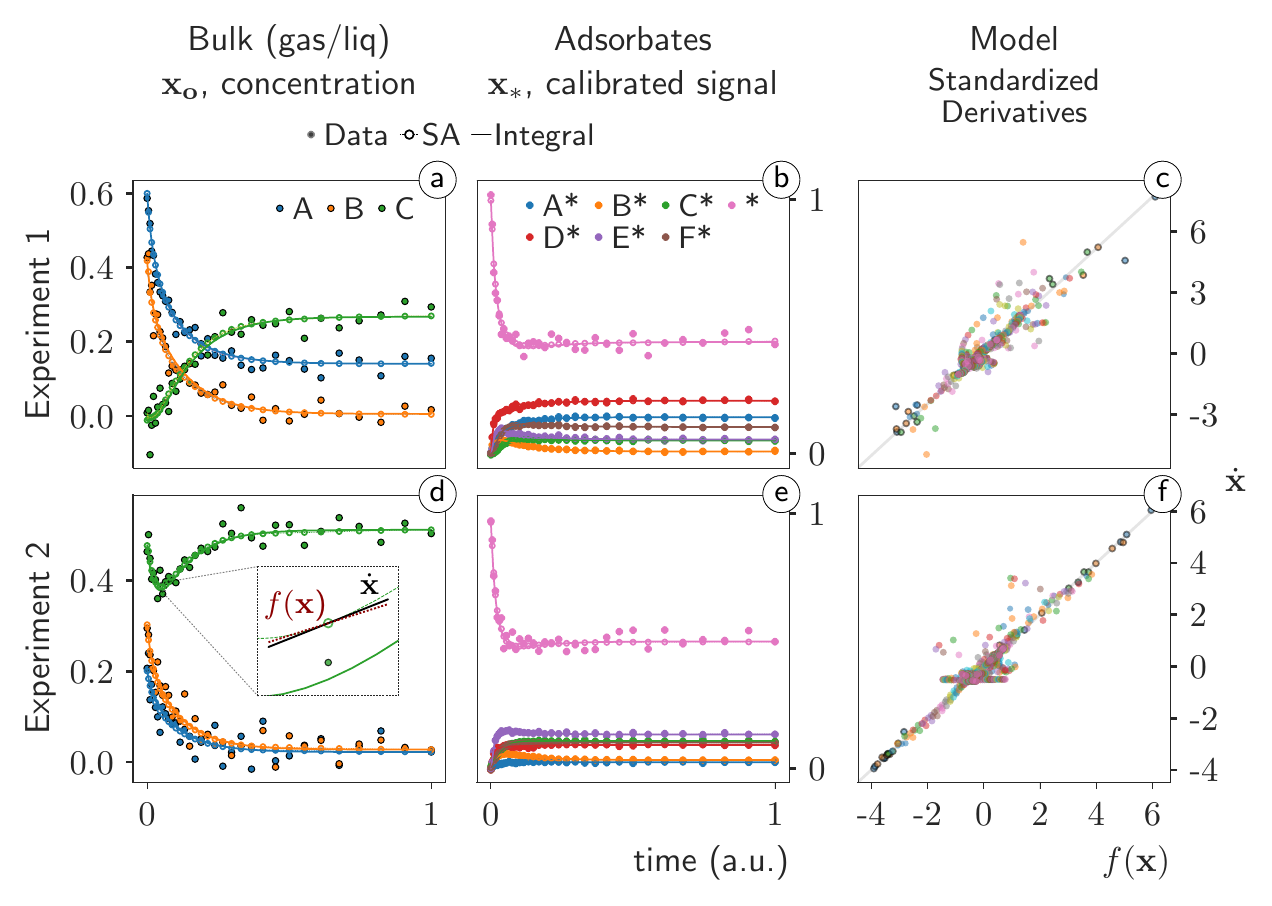}%wid\thetah=1.\linewid\thetah
    \caption{r\kinn{s} results: synthetic data (solid markers), surrogate approximator (open markers) down-sampled to every two datapoints, and numerical integral with initial conditions estimated from SA at $t=0$ for bulk-phase (a and b) and adsorbates (b and e). Parity plot of standardized derivatives for all chemical species (c and f) from the kinetic model $f(\xvec)$ and from algorithmic differentiation of the surrogate approximator ($\dxvec$). Physical model and SA derivatives were standardized based on the mean standard deviation per state to facilitate the visualization of their correlation.}\label{fig:mle_results}
\end{figure}

In \cref{fig:mle_results}, synthetic data (solid circles), SA estimates (open circles, dashed line) and physical model integral based on the initial conditions estimated from SA (solid lines) are plotted for bulk species (a and d) and adsorbates (b and e). Additionally, the physical model, $f(\xvec)$, and SA algorithmic differentiation, $\dxvec$, parity plots for all species are shown in \cref{fig:mle_results} (c) and (f). The results illustrate a strong agreement between the numerical integration of the model with recovered parameters (solid lines) and the synthetic experimental data for both experiments. 
%The variation in $f(\xvec)$ for nearly constant $\dxvec$ near the origin for certain species is mainly an artifact of standardization of concentrations that rapidly reach steady state and, thus, have near-zero standard deviation. 
% Importantly, for SA consisting of deep underlying neural networks, model residuals may be evaluated or sampled at points (independent variable, e.g. time or space) between those where data has been measured. Intermediate sampling hence enforces that the physical model is satisfied at points where data is being interpolated or extrapolated. While on one hand such a procedure can prevent SA from create kinks between available points, on the other hand it may lead to Hessian-based estimates for model parameters confidence intervals be overestimated, i.e., shrink, since the evaluation of model residuals at SA interpolated points introduce inherent bias to the MLE formulation.
%
\section{Discussion}
In this work, we introduce a hyperparameter-free approach for inverse PINNs problems applied to high-dimensional coupled non-linear ODEs. We express the probability density function for the residuals of PINNs applied to heterogeneous chemical kinetics systems, \kinn{s}~\cite{Gusmao2022Kinetics-informedNetworksb}, as normal distributions. We reformulate the \kinn{s} regularized MSE cost functions as MLEs with direct propagation of SA interpolation error through the physical (kinetic) model gradients. The model gradients act as projectors to connect the cost functions for data interpolation and physical derivatives to eliminate the hyperparameter that is often required in the in inverse PINNs.
%the MLE-like cost function from interpolation and physics in such a way we supplant the need for hyperparameter tuning in inverse PINNs given a known physical model structure. 
In addition, we address the issue of ill-conditioning of covariance matrices by using SVD of the coupling matrix to find uncorrelated subspaces that function as pre-conditioners over which the residuals and covariance matrices may be projected. SVD also allows the decomposition of the SA representation into the range (time variant) and nullspace (time invariant) components of the stoichiometry matrix, thus leading to an intrinsic dimensionality reduction at the representation level, where only the range components are mapped by SAs' NNs. Such a decomposition also allows a closed-form solution for the calibration factors needed to relate semi-quantitative signals of latent adsorbate populations to explicit adsorbate concentrations. 
%With known calibration factors, the adsorbate concentration dynamics can be reconstructed, and the nullspace equations can be solved a priori. 

%A flexible SA structure is also presented in \cref{subsec:rkin_mle}, \cref{eq:inv_null_sol}, which provides a scaffold for determining potential surface concentration dynamics in case of partial information. Importantly, any SA solution satisfies the DAE normalization constraints due to the reverse application of the $\opnc$ operator and, hence, the conservation law determined by the material balance. However, in such a scenario, there could possibly exist multiple solutions for the nullspace equations given that multiple calibration factors and associated latent dynamics would satisfy it. 

The novel rKINNs framework with MLE-estimators and the range-nullspace decomposition is applied to a complex case study in heterogeneous catalysis. It efficiently converges to a well-defined solution that is consistent with the results of expensive multi-objective optimization with the prior näive-\kinn{} approach. Moreover, the rKINN formalism allows for rigorous propagation of error between the model and parameter spaces, enabling calculation of error bars on parameter estimates that are robust to noise in the data and the architecture of the underlying NNs 
{The direct connection between interpolation and physics done through the construction of ``physically-coupled'' residuals probability density functions. This approach provides a general self-regularized framework for inverse PINNs problems based on error propagation through MLE forms, which are applicable to high-dimensional coupled ODEs. Even in the case of highly correlated outputs, SVD -- or other orthogonal decomposition methods -- can be used to define orthogonal minimal subspaces for the residuals projection and SA representation and robust evaluation of MLEs as NNs weights are trained.} {This framework applications span other types of reactor models, such as: homogeneous continuous flow stirred-tank reactor equations would include information about the flows at the boundary in the nullspace and, hence, may not be time invariant yet still separable from the kinetics, and uniform plug-flow reactors differential equations, which requires a change in variable over which it is integrated. Finally, additional effort would be necessary to expand the application of the concepts here devised to non-uniform transient systems, where partial differential equations would need to be solved.} 

One limitation of the current rKINN approach is that semi-quantitative signals for all latent species are required. However, it common in heterogeneous catalysis the lack of information about some or all latent species. Additional work is required to address this scenario, where some assumptions about the covariance structure of unobservable species will be required. This is complicated by the fact that the uniqueness of the solution to the inverse problem with incomplete information is not guaranteed, and further studies are needed to establish the limits of solutions obtained from inverse problems in the limit of missing information or extreme stiffness. Moreover, in the case study presented it is assumed the physical model structure is fully known, which is often not the case in heterogeneous catalysis. This can potentially be overcome by coupling mechanism discovery algorithms~\cite{Wu2022PolyODENet:Data, Ji2021AutonomousNetwork,Ji2022AutonomousNetworks} with (r)KINNs or other frameworks for solving the inverse problem, or by the use of NODEs that can learn surrogate functions for the system dynamics~\cite{Rackauckas2020UniversalLearning,Bradley2022TrainingEquations,Yin2023GeneralizedInterpretability,Doppel2022EfficientSteps}. 
%Additional work would be required to extend the calibration factor closed form solution to account for additional degrees of freedom in a scenario of partially available information. 
%TC:ignore
\section{Methods}
\subsection{Probability density function}
\label{subsec:pdf_assumptions}
{We assume that variations about $\xvec$, and incumbent parameters, $\pvec$, in \cref{eq:linearized} are random variables that follow independent multivariate normal distributions:}
 \begin{equation}
\begin{aligned}
    \delta\xvec\in\hat{\erresp}_{\xvec} &\sim\mathcal{N}(0,\cov_{\xvec}) \\
    \delta\pvec\in\hat{\erresp}_{\pvec} &\sim\mathcal{N}(0,\cov_{\pvec})\\
\end{aligned}
\label{eq:epshatxp}
\end{equation}
With these assumptions, we can show that the variations about $\hat{\erresp}_{\dxvec_i}$ also comprise a multivariate normal distribution, whose covariance matrix can be estimated by the outer product of \cref{eq:linearized} at timepoint $i$. Neglecting $\mathcal{O}(\erresp^2)$ terms and assuming orthogonality between $\hat{\erresp}_\xvec$ and $\hat{\erresp}_\pvec$ yields:
\begin{equation}
\begin{aligned}
\cov_{\dxvec_i} &= \{\cov_{\dxvec_i}|\xvec_i,\dxvec_i,\pvec\}=\langle\delta{\erresp}_{\dxvec_i}\otimes\delta{\erresp}_{\dxvec_i}\rangle\\&= \langle(-\partial_\xvec f_i\hat{\erresp}_\xvec-\partial_\pvec f_i\hat{\erresp}_\pvec)\otimes(-\partial_\xvec f_i\hat{\erresp}_\xvec-\partial_\pvec f_i\hat{\erresp}_\pvec)\rangle\\
        &= \partial_\xvec f_i\langle \hat{\erresp}_\xvec\otimes \hat{\erresp}_\xvec\rangle \left(\partial_\xvec f_i\right)^T+\partial_\pvec f_i\langle \hat{\erresp}_\pvec\otimes \hat{\erresp}_\pvec\rangle 
        \left(\partial_\pvec f_i\right)^T\text{ for }\hat{\erresp}_\xvec\perp\hat{\erresp}_\pvec\\
        &=\partial_\xvec f_i\cov_\xvec \left(\partial_\xvec f_i\right)^T+\partial_\pvec f_i\cov_\pvec\left(\partial_\pvec f_i\right)^T\\
        &=\cov^\xvec_{\dxvec_i}+\cov^{\pvec}_{\dxvec_i}\label{eq:sigmaouter}
\end{aligned}
\end{equation}
where the arguments of the physical model $f$ are suppressed for brevity, and $f_i=f(\xvec_i,\pvec)$. This implies that the perturbations to the model residual can also be considered as a random variable that follows a multivariate normal distribution:
\begin{equation}
\delta\erresp_{\dxvec_i} \in \hat{\erresp}_{\dxvec_i}\sim\mathcal{N}(0,\cov_{\dxvec_i})
\label{eq:epshatxdot}
\end{equation}
Notably, the covariance matrix for the underlying model parameter error distribution is generally unavailable, since the true parameters, $\pvec$, are unknown. However, \cref{eq:linearized} can be re-arranged to provide a least-squares approximation for $\erresp_\pvec$, where the current evaluated states and derivative errors, $\erresp_\xvec$ and $\erresp_\dxvec$ are considered as samples drawn from the underlying random variables defined in \cref{eq:epshatxp} and \cref{eq:epshatxdot}, and the representation contribution in $\partial_\xvec\dxvec$ is neglected since model parameter errors should not depend on the SA: 
\begin{align}
    {\erresp}_{\pvec_i}&=\left(\partial_{\pvec}f_i^T\partial_{\pvec}f_i\right)^{-1}\partial_{\pvec}f_i^T\left({{\erresp}}_{\dxvec_i}-\partial_\xvec f_i{{\erresp}}_{\xvec_i}\right)\label{eq:epsp_err}
\end{align}
We assume that the covariance of the data interpolation can be obtained from the residuals in $
\xvec$. Letting $\erresp_{\xvec}$ be a sample of the random variable $\hat{\erresp}_{\xvec}$ (where $\erresp_{\xvec}$ is defined by \cref{eq:res_interp}) yields:
\begin{equation}
\cov_\xvec = \langle\erresp_\xvec-\langle\erresp_\xvec\rangle\otimes\erresp_\xvec-\langle\erresp_\xvec\rangle\rangle \label{eq:sigma_x}   
\end{equation}
In this case, ``recentering'' (subtracting the sample mean when estimating sample covariances) is not strictly necessary; however, to avoid numerical instabilities from large initial residuals, intermediate estimated covariances are recentered during the \kinn{s} training, which does not affect the final results given that the fundamental structural assumption is that $\langle\erresp_\xvec\rangle\to\mathbf{0}$ as the optimization proceeds. Notably, \cref{eq:epsp_err} yields a time-dependent parameter error, ${\erresp}_{\pvec_i}$, based on the time-dependence of $\erresp_\xvec$ and $\erresp_\dxvec$, and is hence different from the true time-independent parameter error, ${\erresp}_{\pvec}^*$ (\cref{eq:err_pvec}). However, since we are treating $\erresp_\xvec$ and $\erresp_\dxvec$ as independent samples from underlying multivariate normal distributions, the interpretation is that ${\erresp}_{\pvec_i}$ is also a sample from $\hat{\erresp}_{\pvec}$ (\cref{eq:epshatxp}) that obeys the structure imposed by \cref{eq:epsp_err}. Thus, the model parameter covariance matrix can be estimated as $\cov_{\pvec} = \langle {\erresp}_{\pvec} - \langle {\erresp}_{\pvec}\rangle  \otimes {\erresp}_{\pvec}-\langle {\erresp}_{\pvec}\rangle \rangle$, with $\langle \erresp_\pvec\rangle\to\mathbf{0}$ as the optimization reaches a critical point. 
\subsection{Covariance inversion stabilization}
In the MLE formulation, a point of attention is that often the data- and model-parameter covariance structures are unknown, either requiring a priori assumptions or the utilization of methods for covariance inference from the data itself~\cite{Rajamani2009EstimationWeighting,Roelant2007NoiseResponses} or from sample residuals~\cite{Rousseeuw1999AEstimator,Ledoit2004AMatrices,Chen2010ShrinkageEstimation}. In this work, we assume a lack of prior knowledge, and thus that the interpolation error is homoscedastic. The precision matrix is then updated periodically throughout the optimization at each epoch. 
To stabilize successive covariance inversions and account for the fact that residuals are not initially centered, the mean squared error is added as uncorrelated additive Gaussian noise to the covariances, \cref{eq:err_stabilization}, which is similar to the strategy adopted to stabilize Gaussian Process Regression~\cite{Ambikasaran2016FastProcesses}. As $\langle \erresp_\xvec\rangle\to0$ at the critical points of $\lt$, this approach will not affect the final solution of the inverse problem.
\begin{equation}
\begin{aligned}
    \cov_{\xvec/\dxvec}&\xleftarrow{}\cov_{\xvec/\dxvec}+\operatorname{diag}(|\langle{\erresp}_\xvec\rangle|)\label{eq:err_stabilization}
\end{aligned}    
\end{equation}
As the model converges this stabilization term reduces, and can be removed prior to error estimation to provide more accurate error estimates. 

\subsection{Neural-network representation of heterogeneous systems}
\label{subsec:nn_representation}
Heterogeneous chemical systems comprise unbound (gas-phase), adsorbed molecules and intermediates. Unbound species (observable) and adsorbate or intermediate -- latent -- species are represented by individual NNs $\mathbf{\hat{x}_{o}}$ and $\mathbf{\hat{x}_\ls}$, respectively, where a normalization operator $\opnc$ was introduced to enforce normalization of surface species (DAE constraint) by construction, i.e., $0\le\mathbf{x_\ls}=\opnc[\mathbf{\hat{x}_\ls}]\le1$ and $|\mathbf{x_\ls}|_1=1\;\forall\;\mathbf{\hat{x}_\ls}\in\mathbb{R}^{\operatorname{dim}(\mathbf{{x}_\ls)-1}}$. In this work, we have reformulated the normalization operator $\opnc$ from its previous trigonometric description to a simpler structure in terms of products of logistic functions that enforces bijection (\SA, \ref{supp:sec:normalization_constraints}).
\begin{align}
    \mathbf{x}&=\begin{bmatrix}\mathbf{\hat{x}_{o}}\\\opnc[\mathbf{\hat{x}_\ls}]\end{bmatrix}\label{eq:ode_op_c}
\end{align}
\subsection{KINNs architecture}
In this work, derivatives are obtained through algorithmic differentiation (AD) with just-in-time (JIT) compilation using the JAX library~\cite{47008,jax2018github}. In addition, SA training and model fitting are trained using the \emph{Adam} optimization algorithm. The code necessary for reproducing the results of this work can be found on GitHub (\url{https://github.com/gusmaogabriels/kinn/tree/rkinns}). 

The MLE-formulation was utilized to build the cost function to be minimized for learning the ODE solutions using r{\kinn}s. The JAX library provided the routines for the forward AD, allowing the obtainment of the gradients with respect to the SAs' parameters, which were optimized with the \emph{Adam} algorithm~\cite{Kingma2015Adam:Optimization}. For both \kinn{s} and r\kinn{s} the underlying SA NNs consists of three-hidden layers, each of which with $20$ neurons $\{20_{\times3}\}$ and corresponding activation functions $\{tanh,\,swish,\,tanh\}$. 

\subsection{Data Generation}
Two sets of initial conditions (\ic{s}) were used to generate synthetic data, which were built to represent scenarios of initially high and low concentration of reactants and products: \ic 1, $\{x_A,x_B,x_C,x_*\}_{t_0}=\{0.6,0.4,0.0,1.0\}$, and \ic 2, $\{x_A,x_B,x_C,x_*\}_{t_0}=\{0.2,0.3,0.5,1.0\}$, respectively. Additive homoscedastic noise with $\sigma = 0.025$ was added to the generated data. Signals for surface coverages were re-normalized to simulate unknown calibration factors, although it is assumed that semi-quantitative signals for all surface states are known.
Numerical integrations for synthetic data generation were carried out with the stiff-nonstiff algorithm LSODA from the FORTRAN ODEPACK~\cite{Hindmarsh1980LSODESolvers,Petzold1983AutomaticEquations}. Forward and reverse rate constants conveyed in \cref{tab:rate_cnstnt} were chosen to depict over two orders-of-magnitude variance in rate constants that typical of reaction networks in heterogeneous catalysis. A total of $100$ time-points were sampled in logarithmic-space, i.e., more points were sampled at the beginning of the experiments, where derivatives exhibit higher variance.
\begin{table}[hbt!]
    \centering
    \caption{Forward and reverse reactions rate constants for the complex reaction mechanism (\textit{dcs}) synthetics data generation.}
    \label{tab:rate_cnstnt}
\begin{tabular}{lcccccccc}
\toprule
{}&\multicolumn{7}{c}{Rate Constants}\\\cline{2-8}\\[-2.05ex]
Type&$\frac{k_{1}^d}{k_{\text{-}1}^d}$&$\frac{k_{2}^d}{k_{\text{-}2}^d}$&$\frac{k_{3}^d}{k_{\text{-}3}^d}$&$\frac{k_{3}^c}{k_{\text{-}3}^c}$&$\frac{k_{4}^c}{k_{\text{-}4}^c}$&$\frac{k_{5}^c}{k_{\text{-}5}^c}$&$\frac{k_{1}^s}{k_{\text{-}1}^s}$\\
\midrule
{forward}&${20}$&${24}$&${16}$&${640}$&${160}$&${560}$&$640$\\
{reverse}&${8}$&${12}$&${40}$&${960}$&${80}$&${160}$&${240}$\\
\bottomrule
\end{tabular}
\end{table}
\subsection{Latent states}
In many complex dynamical systems the species can be decomposed into measurable states and ``latent'' states that cannot be directly observed. This is the case in heterogeneous catalysis, where unbound gas phase species are typically observable, and denoted by $\cvec_\mathbf{o}$, while $\cvec_\ls$ denotes the concentration of the latent variables, i.e., adsorbed and intermediate surface species.  In this case, the range and null-space basis can be distinguished as:
\begin{equation}    \mathbf{U}=\begin{bmatrix}\begin{bmatrix}\mathbf{U}_\mathbf{o}\\\mathbf{U}_\ls\end{bmatrix}^{\sR}&\begin{bmatrix}\mathbf{U}_\mathbf{o}\\\mathbf{U}_\ls\end{bmatrix}^{\sN}\end{bmatrix}\label{eq:de_reac_red_gl}
\end{equation}
This leads to partitioned concentration vectors in \cref{eq:conc_rn_gl}.
\begin{align}
    \begin{bmatrix}\cvec_\mathbf{o}\\\cvec_\ls\end{bmatrix}_{(t)}&=\begin{bmatrix}\cvec_\mathbf{o}\\\cvec_\ls\end{bmatrix}^{\sR}_{(t)}+\begin{bmatrix}\cvec_\mathbf{o}\\\cvec_\ls\end{bmatrix}^{\sN}\label{eq:conc_rn_gl}
\end{align}
Latent dynamics are characterized by incomplete information about the evolution of a subset of states, which can be partially ``hidden'' or indirectly inferable through the calibration of intensity measurements in conjunction with known associated conservation laws. For example, in heterogeneous catalysis, operando analytical methods, such as temperature-programmed reaction spectroscopy, may be utilized to track catalyst surface composition over time-on-stream~\cite{Madix1978}. The raw spectroscopic data (signal), $\mathbf{y}$ is linear correlated with the latent states $\cvec_\ls$, as in \cref{eq:calib}, with calibration factors, $\boldsymbol{\gamma}$, based on the conservation of total active sites constraints. 
\begin{equation}
    \begin{aligned}
        \cvec_\ls=\yvec_\ls\circ\boldsymbol{\gamma},\quad\text{s.t.}\,\mathbf{1}^T\cvec_\ls=1
    \end{aligned}\label{eq:calib}
\end{equation}
In \cref{supp:sec:data_recnstrctn}, we use the null-space equations to a derive a closed-form solution for the calibration factor $\boldsymbol{\gamma}$ in the scenario where $\mathbf{y}_\ls$ signal can be retrieved.

\subsection{{Robust KINNs - MLE \& Range-Nullspace decomposition}}
\label{subsec:rkin_mle} 
\begin{figure}[hbt!]
    \centering
    \includegraphics[keepaspectratio=true,scale=1.,clip=True,page=2]{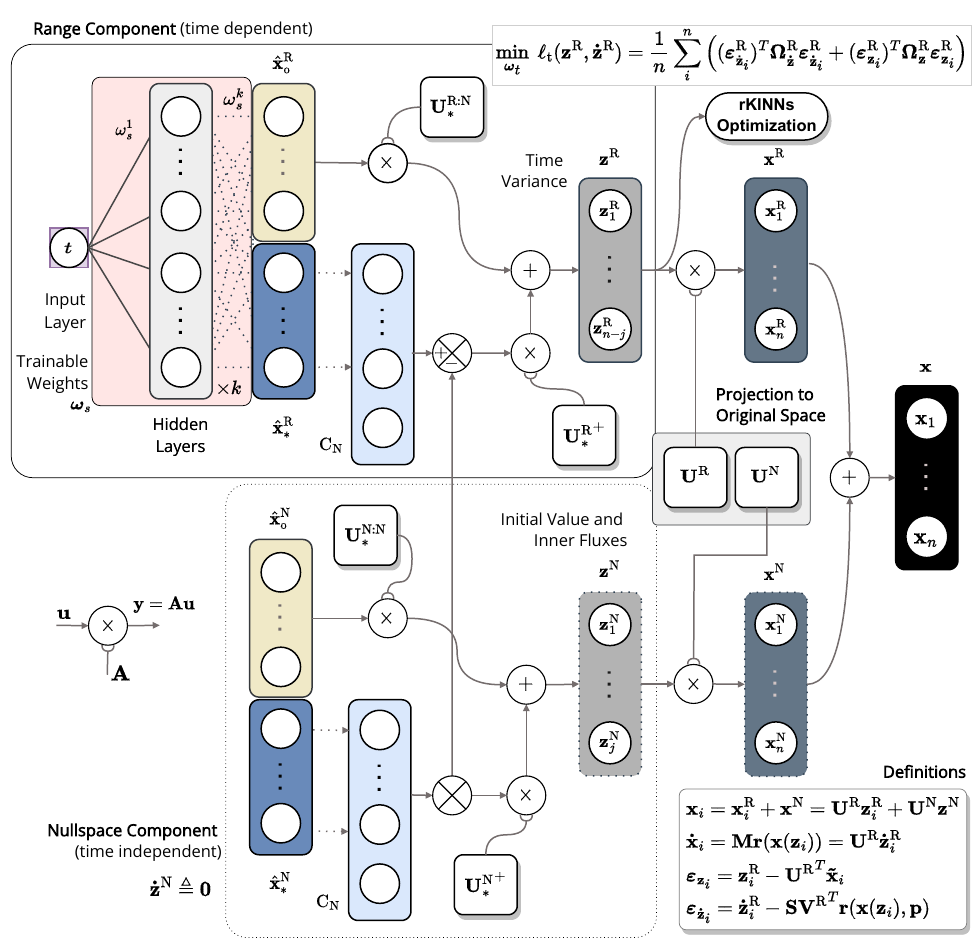}
    \caption{Surrogate approximator structure based on range-nullspace decomposition using SVD. Initial-values and inner fluxes are contained in the time-invariant nullspace component (bottom). Time variations are represented by the fully-connected feed-forward neural network (top) with $k$ hidden layers, which includes both unbound, $\mathbf{\hat{\xvec}_{o}}^\sR$, and bound, ${\mathbf{\hat{\xvec}_{\ls}}}^{\sR}$ latent representations that, through $\opnc$ and linear operations, leads to $\zvec^\sR$. The MLE framework is constructed around the $\mathbf{z}^\sR$ lower dimension representation assuming that $\mathbf{z}^\sN$ can be obtained directly from measured data.}\label{fig:sa_structure}
\end{figure}

In \cref{eq:kin_ode}, the underlying ODE can be decomposed into its components in the nullspace ($\sN(\mathbf{M})$) and range ($\sR(\mathbf{M})$), and SVD provides a basis for the representation of states and its derivatives in those spaces. Estimated variances can be propagated and separated into nullspace and range components. Such a decomposition eliminates linearly-dependent terms, creates denser covariance matrices, and mitigates the propagation of numerical errors to the precision matrices due to ill-conditioned covariance matrix-inversions. {The SA representation is constructed to structurally separate the orthogonal components of the time-independent (nullspace) and -dependent (range) solutions, \cref{fig:sa_structure}. By combining \cref{eq:conc_rn,eq:crange,eq:cnull} and \cref{eq:de_reac_red_gl}, the SA is built as a function of the nullspace and range coverage solutions, i.e. $\opnc[\cdot]$ in \cref{eq:inv_null_sol} and \cref{eq:inv_range_sol}, respectively. The time-invariant solution $\zvec^\sN$ consists of the sum of a solution in the convex-hull of $(\mathbf{U}^\sN_\ls)^T$, since $\opnc[\mathbf{\hat{x}_\ls^\sN}]$ implies a convex-affine combination of its basis vectors, translated by solutions in the positive linear-span of $(\mathbf{U}^\sN_\mathbf{o})^T$, which parameterized by $\mathbf{\hat{x}_{o}^\sN}$ through $\mathbf{U}_\ls^{\sN:\sN}$. A visual representation of how the all subspaces are interconnected is shown in \cref{fig:sa_structure}}.
\begin{align}
        \zvec^\sN&=\left(\mathbf{U}^\sN_\ls\right)^{+}\opnc[\mathbf{\hat{x}_\ls^\sN}]+\mathbf{U}^{\sN:\sN}_\ls\mathbf{\hat{x}_{o}^\sN}\label{eq:inv_null_sol}
\end{align}
The range-component solution depends on the nullspace estimates for the latent variables, $\xvec^\sN_\ls=\mathbf{U}^\sN_\ls\zvec^\sN$. Similarly to the nullspace solution, to enforce normalization, $\xvec_\ls=\opnc[{\mathbf{\hat{x}_\ls^\sR}}{}_{(t)}]$ is first calculated and variations in the latent species solutions $\zvec^\sR$ are parameterized by $\mathbf{U}^{\sR:\sN}_\ls\in\mathrm{N}[\mathbf{U}^\sR_\ls]$, as in \cref{eq:inv_range_sol}. Importantly, components in $\mathbf{U}^{\sN:\sN}_\ls$ and $\mathbf{U}^{\sR:\sN}_\ls$ are orthogonal to the latent space solution and, therefore, represent variations in the gas phase that do not affect the dynamics of the surface phase. The latent space solution is computed as
\begin{align}
    \zvec^\sR&=\left(\mathbf{U}^\sR_\ls\right)^{+}\left(\opnc[\mathbf{\hat{x}_\ls^\sR}{}_{(t)}]-\xvec^\sN_\ls\right)+\mathbf{U}^{\sR:\sN}_{\ls}\mathbf{\hat{x}_{o}}^\sR{}_{(t)}\label{eq:inv_range_sol}
\end{align}
where $\mathbf{\hat{x}_{o}^\sR{}_{(t)}}\in\mathbb{R}^d$ for $d=\operatorname{nullity}[\mathbf{U}^\sR_\ls]$, and $\mathbf{\hat{x}_{o}^\sN{}_{(t)}}\in\mathbb{R}^d$ for $d=\operatorname{nullity}[\mathbf{U}^\sN_\ls]$. 
~~
With such a structure, all the variance is constrained to the $\sR[\mathbf{M}]$, which is parameterized by the time-invariant nullspace. We refer to the reduced optimization problem formulation as r\kinn{} 
 (Robust KINN), which is conveyed in \cref{eq:loglikhd_rn}, where $\varepsilon_{\zvec}^\sR=(\mathbf{U}^\sR)^T\tilde{\xvec}-\zvec^\sR=(\mathbf{U}^\sR)^T\varepsilon_{\xvec}$  and $\varepsilon_{\dzvec}^\sR=(\mathbf{U}^\sR)^Tf(\xvec(\zvec),\pvec)-\dzvec^\sR=(\mathbf{U}^\sR)^T\varepsilon_{\dxvec}$ are the state- and state-derivative errors projected onto the $\sR[\mathbf{M}]$ subspace, respectively. This leads to a revised definition of the optimization problem,
\begin{equation}
\begin{aligned}
    \min_{\parlumped} \quad &\lt=\frac{1}{n}\sum_i^n\left((\erresp_{\dot{\zvec}_i}^\sR)^T\icov_{\dzvec_i}^\sR\erresp_{\dzvec_i}^\sR+(\erresp_{\zvec_i}^\sR)^T\icov_{\zvec}^\sR\erresp_{\zvec_i}^\sR\right) \\
    \textrm{s.t.}\quad&\icov_{\dzvec_i}^\mathrm{\sR}=\left((\mathbf{U}^{\sR})^T\left(\cov^\xvec_{\dxvec_i}+\cov^{\pvec}_{\dxvec_i}\right)\mathbf{U}^{\sR}\right)^{-1}\\
    \quad&\icov_\zvec^{\sR}=\left((\mathbf{U}^{\sR})^T\cov_{\xvec}\mathbf{U}^{\sR}\right)^{-1}\\
    \quad&\pvec\in\mathbb{R}^{\operatorname{dim}(\pvec)},\;\parsm\in\mathbb{R}^{\operatorname{dim}(\parsm)}
    \label{eq:loglikhd_rn}
\end{aligned}    
\end{equation}
which is solved with the rKINNs \cref{alg:kinns_inv_mle}.

~\begin{algorithm}
\caption{Robust \kinn{s}}\label{alg:kinns_inv_mle}
\begin{algorithmic}[2]
        \State \algorithmicrequire{ Observed data ($\mathbf{t}$, $\tilde{\xvec}$), stoichiometry matrix $\mathbf{M}$, max. iterations $n_{max}$, no. of epochs $n_e$, tolerance $tol$}
        \State Build NN, initialize SA parameters $\parsm$, kinetic model parameters $\pvec$
        \State Evaluate $\icov_{\zvec}^{\sR}$ and $\icov_{\dzvec}^{\sR}$.
        \While{epoch < $n_e$  \textbf{or} $\lt(\parlumped)$ < $tol$}   
        \While{iter < $n_{max}$}
        \State Predict states $\xvec(t,\parsm)$ from SA over $\mathbf{t}$ 
        \State Predict states derivatives $\dxvec(t,\parsm)$ from SA over $\mathbf{t}$ 
        \State Compute projected errors $\boldsymbol{\varepsilon}^\sR_\zvec$ and $\boldsymbol{\varepsilon}^\sR_\dzvec$.
        \State Compute objective function {$\lt(\parlumped)$} over $\mathbf{t}$
        \State Compute gradients of {$\lt(\parlumped)$} with AD; $\partial_{\parlumped}\lt(\parsm,\pvec)$
        \State Update SA parameters $\parsm$ and kinetic model parameters $\pvec$ using \emph{Adam}
        \EndWhile
        \State Update $\partial_\xvec f(\xvec,\pvec)$ and $\partial_\pvec f(\xvec,\pvec)$
        \State Update $\cov^\xvec_{\dxvec_i}$ and $\cov^{\pvec}_{\dxvec_i}$
        \State Update $\cov_\xvec$ and $\cov_\dxvec$ with uncorrelated additive noise $|\langle{\erresp}_{\xvec}\rangle|$
        \State Update covariance matrices $\cov_{\xvec}^\sR$ and $\cov_{\dxvec_i}^\sR$
        % with boosting at $\beta$
        \State Evaluate precision matrices $\icov_{\zvec}^\sR$ and $\icov_{\dzvec_i}^\sR$ 
        \EndWhile
        \State \algorithmicensure{ Kinetic model parameters $\pvec$}
\end{algorithmic}
\end{algorithm}~

\section*{Acknowledgements}
G.S.G. is a PhD research fellow with International Business Machine Corporation (IBM) since mid-2021 through the IBM Global University Program Awards. Acknowledgement is made to the donors of the American Chemical Society Petroleum Research Fund for partial support of this research. This work was supported by the U.S. Department of Energy (USDOE), Office of Energy Efficiency and Renewable Energy (EERE), Industrial Efficiency and Decarbonization Office (IEDO) Next Generation R\&D Project DE-FOA-0002252-1775 under contract no. DE-AC07-05ID14517. We acknowledge Dr. John Kitchin for his formative work on the utilization of neural networks as bases for the solution of simple coupled forward kinetics ODEs (\url{kitchingroup.cheme.cmu.edu/blog}). The authors are grateful to Dr. M. Ross Kunz and Dr. Adam Yonge for the enlightening discussions involving the maximum-likelihood formulation of PINNs problems and uncertainty quantification, to Ms. Sushree Jagriti Sahoo for her comments on the algebraic decomposition of ODE coupling matrices, and to Mr. Dingqi Nai for general assessments of the application of KINNs to experimental data. G.S.G. thanks especially Dr. William Bradley and Mr. Zachary Kilwein for the extensive debate-like discussions on the application of PINNs and NODEs to general Chemical Engineering problems.

\section*{Conflict of Interest}
The authors declare the nonexistence of conflicts of interest.

\printnomenclature
%TC:endignore
\bibliography{main}
\bibliographystyle{unsrt} % GSG change unsrt
\end{document}

\maketitle

\section{Normalization DAE Constraints Revisited}\label{sec:normalization_constraints}

The formulation of the structural or topological normalization DAE constraints in the naïve \kinn{s} work~\protect{\cite{Gusmao2022Kinetics-informedNetworksb}} was inspired by trigonometric relations or Pythagorean theorem outcomes for radius-$1$ hypersphere. Originally in \kinn{s}, we had $\mathbf{x}(\parsm,t)\in\mathbb{R}^k$ be the output values of the normalization constraint operator with underlying neural network $\hat{x}(\parsm,t)\in\mathbb{R}^{k-1}$, such that
\begin{align}\label{eqn:normconstr}
\begin{split}
    {x}_i(\parsm,t)&=\left(1-\sin^{2}\left({\hat{x}}_i(\parsm,t)\right)\right)\prod_{j<i}\sin^{2}\left({\hat{x}}_j(\parsm,t)\right)\;\forall\;i<p;\;{i,j}\in\mathbb{N}\\
    {x}_p(\parsm,t)&=\prod_{j<p}\sin^{2}\left({\hat{x}}_j(\parsm,t)\right)
\end{split}
\end{align}
Where $\mathbf{x}=\mathrm{C_N}[\mathbf{\hat{x}}]$, such that $0\le\mathbf{x}\le1$ and $\sum\mathbf{x}=1\;\forall\;\mathbf{\hat{x}}\in\mathbb{R}^{p-1}$, which consists of a DAE-constraint since $\sum \dot{\mathbf{x}}=0$. A problematic property of $\mathrm{C_N}$ representation as a function of $\sin^2(\cdot)$ arises from its non-injective property due to its periodicity in values, i.e., $\sin^2(\alpha)=\sin^2(\alpha+n\pi)$, and correspondence in its derivatives, i.e., $$\partial_\alpha\sin^2(\alpha)=\partial_\alpha\sin^2(\alpha+n\pi),\forall n\in\mathbb{Z},\,\alpha\in\mathbb{R}.$$ If we let $\hat{x}(t,\parsm)=\alpha$ and $\hat{x}(t,\parsm+\delta\parsm)=\alpha+n\pi$, the derivatives values depend on the structure of $\hat{x}$, i.e., with $\sin^2(\hat{x}(t,\parsm))=\sin^2(\hat{x}(t,\parsm+\delta\parsm))$, we have 
\begin{align*}
\partial_t\sin^2(\hat{x}(t,\parsm))&=\partial_{\hat{x}}\sin^2(\hat{x}(t,\parsm))\partial_t\hat{x}(t,\parsm)\\&=\partial_{\hat{x}}\sin^2(\hat{x}(t,\parsm+\delta\parsm))\partial_t\hat{x}(t,\parsm+\delta\parsm)\iff\partial_t\hat{x}(t,\parsm)=\partial_t\hat{x}(t,\parsm+\delta\parsm)    
\end{align*}
for a $\delta\parsm$ variation in $\parsm$. 

For ``overly complex'' neural networks and noisy data, changes in derivatives sign during training may drive the inverse \kinn{s} SA to get trapped into local minima. That because SA estimates may be consistent with the measured data (interpolation) but the automatic differentiation of the SA has the opposite sign, thus leading the physical model parameters to shrink down to machine precision. In summary, in such a scenario the inverse \kinn{s} training halts without an obvious reason requiring further detailed inspection to trace the numerical inconsistency. To circumvent such a scenario, we propose replacing the trigonometric formulation built as of $\sin^2(\cdot)$ with the sigmoid (or logistic) function $\operatorname{\sigma}(\cdot)$, which is injective and surjective (bujective), is also bounded between $0$ and $1$ and satisfies all the properties required for the DAE constraint operator.

\begin{align}\label{eqn:normconstr}
\begin{split}
    {x}_i(\parsm,t)&=\left(1-\operatorname{\sigma}\left({\hat{x}}_i(\parsm,t)\right)\right)\prod_{j<i}\operatorname{\sigma}\left({\hat{x}}_j(\parsm,t)\right)\;\forall\;i<p;\;{i,j}\in\mathbb{N}\\
    {x}_p(\parsm,t)&=\prod_{j<p}\operatorname{\sigma}\left({\hat{x}}_j(\parsm,t)\right)
\end{split}
\end{align}

\section{MLE Optimality Conditions}
The optimality condition for cost function of the minimization problem in \cref{main:eq:loglikhd} can be represented as expected values in \cref{eq:loglikhdexp}.
\begin{equation}
\nabla\left\langle\erresp_{\dot{\xvec}}^T\icov_{\dxvec}\erresp_{\dxvec}+\erresp_{\xvec}^T\icov_{\xvec}\erresp_{\xvec}\right\rangle=\mathbf{0}\label{eq:loglikhdexp}
\end{equation}
Which can be expanded over the independent variables $\xvec$ and $\pvec$ as in
\begin{align}
\nabla\left\langle\erresp_{\dot{\xvec}}^T\icov_{\dxvec}\erresp_{\dxvec}+\erresp_{\xvec}^T\icov_{\xvec}\erresp_{\xvec}\right\rangle&=\left[\partial_\xvec,\partial_\dxvec,\partial_\pvec\right]\left\langle\erresp_{\dot{\xvec}}^T\icov_{\dxvec}\erresp_{\dxvec}+\erresp_{\xvec}^T\icov_{\xvec}\erresp_{\xvec}\right\rangle\\
&\propto\left\langle\left[\partial_\xvec\dxvec+\mathbf{I}-\partial_\xvec f(\xvec,\pvec),\;-\partial_{\pvec}f(\xvec,\pvec)\right]^T\icov_{\dxvec}\erresp_{\dxvec}\right\rangle-\left\langle\left[\partial_\dxvec\xvec,\;\mathbf{I}\right]^T\icov_{\xvec}\erresp_\xvec\right\rangle\\
&=-\left\langle
\begin{bmatrix}
    \partial_\xvec f(\xvec,\pvec)^T-\left(\partial_\xvec\dxvec\right)^T&\mathbf{I}\\
    \partial_{\pvec}f(\xvec,\pvec)^T&\mathbf{0}\\
    -\mathbf{I}&\left(\partial_\dxvec\xvec\right)^T
\end{bmatrix}
\begin{bmatrix}
    \icov_{\dxvec}&\mathbf{0}\\
    \mathbf{0}&\icov_{\xvec}
\end{bmatrix}
\begin{bmatrix}
    \erresp_\dxvec\\\erresp_\xvec
\end{bmatrix}\right\rangle=\mathbf{0}\label{eq:optmatrix}
\end{align}
With the assumption that $\partial_\xvec\dxvec$ and $\partial_\dxvec\xvec=(\partial_\xvec\dxvec)^{-1}$ do not contribute to structural variance, the condition defined in \cref{eq:optmatrix} can be simplified to
\begin{equation}
\left\langle
\begin{bmatrix}
    \partial_\xvec f(\xvec,\pvec)^T&\mathbf{I}\\
    \partial_{\pvec}f(\xvec,\pvec)^T&\mathbf{0}
\end{bmatrix}
\begin{bmatrix}
    \icov_{\dxvec}&\mathbf{0}\\
    \mathbf{0}&\icov_{\xvec}
\end{bmatrix}
\begin{bmatrix}
    \erresp_\dxvec\\\erresp_\xvec
\end{bmatrix}\right\rangle=\mathbf{0}\label{eq:optmatrix_simp}
\end{equation}
From the final results in \cref{main:eq:sigmaouter}, \cref{eq:optmatrix_simp} can be expressed as a function about $\icov_\xvec$ and $\icov_\pvec$ only, but expressing $\partial_\xvec f(\xvec_i,\pvec)$ and $\partial_\pvec f(\xvec_i,\pvec)$ in terms of their singular value decompositions.
\begin{equation}
    \begin{aligned}
        \partial_\xvec f(\xvec,\pvec)&=\mathbf{U}_\xvec\mathbf{S}_\xvec\mathbf{V}_\xvec^T\\
        \partial_\pvec f(\xvec,\pvec)&=\mathbf{U}_\pvec\mathbf{S}_\pvec\mathbf{V}_\pvec^T\\
        \text{ with }\mathbf{U}^T\mathbf{U}&=\mathbf{I},\;\mathbf{V}^T\mathbf{V}=\mathbf{I}\\
        \text{ and }\mathbf{S}&=\operatorname{diag}(\mathbf{s}),\;\text{singular values}
    \end{aligned}
\end{equation}
Which leads to \cref{eq:optmatrix_simp2} for the general case where $\operatorname{dim}(\pvec)>\operatorname{dim}(\xvec)$.
\begin{equation}
\begin{aligned}
\left\langle
\begin{bmatrix}
    \mathbf{V}_\xvec\mathbf{S}_\xvec\mathbf{U}_\xvec^T\mathbf{U}_\xvec\mathbf{S}_\xvec^{-1}\mathbf{V}_\xvec^T\icov_\xvec\mathbf{V}_\xvec\mathbf{S}^{-1}_\xvec\mathbf{U}_\xvec^T&\icov_\xvec\\
    \mathbf{V}_\pvec\mathbf{S}_\pvec\mathbf{U}_\pvec^T\mathbf{U}_\pvec\mathbf{S}_\pvec^{-1}\mathbf{V}_\pvec^T\icov_\pvec\mathbf{V}_\pvec\mathbf{S}^{-1}_\pvec\mathbf{U}_\pvec^T&\mathbf{0}
\end{bmatrix}
\begin{bmatrix}
    \erresp_\dxvec\\\erresp_\xvec
\end{bmatrix}\right\rangle
&=\left\langle
\begin{bmatrix}
    \icov_\xvec\left(\left(\partial_\xvec f(\xvec,\pvec)^T\right)^+\erresp_\dxvec+\erresp_\xvec\right)\\
    \mathbf{V}_\pvec\mathbf{V}_\pvec^T\icov_\pvec\left(\partial_\pvec f(\xvec,\pvec)^T\right)^+\erresp_\dxvec
\end{bmatrix}
\right\rangle \\
&=
\begin{bmatrix}
    \icov_\xvec\langle\left(\partial_\xvec f(\xvec,\pvec)^T\right)^+\erresp_\dxvec+\erresp_\xvec\rangle\\
    \langle\mathbf{V}_\pvec\mathbf{V}_\pvec^T\icov_\pvec \left(\partial_\pvec f(\xvec,\pvec)^T\right)^+\erresp_\dxvec\rangle
\end{bmatrix}=\mathbf{0}
\end{aligned}\label{eq:optmatrix_simp2}
\end{equation}
In which $\mathbf{A}^+$ is the Moore-Penrose pseudo-inverse of matrix $\mathbf{A}$~\protect{\cite{Moore1920,Penrose1955}}. For $\icov_\xvec$ full-rank, the optimality conditions simplify to:
\begin{align}
        \left\langle\erresp_\xvec\right\rangle&=\mathbf{0}\label{eq:condx}\\
        \langle\left(\partial_\xvec f(\xvec,\pvec)^T\right)^+\erresp_\dxvec\rangle&=\mathbf{0}\label{eq:conddx}\\
        \langle\mathbf{V}_\pvec\mathbf{V}_\pvec^T\icov_\pvec \left(\partial_\pvec f(\xvec,\pvec)^T\right)^+\erresp_\dxvec\rangle&=\mathbf{0}\label{eq:condp}
\end{align}
Condition \cref{eq:condx} follows naturally from the assumption of centered interpolation errors. Condition \cref{eq:conddx} conveys the expected value for orthogonality between derivative residuals and the range of $\partial_\xvec f$, and condition \cref{eq:condp} denotes the expected value for the relationship between derivative residuals and gradients of the physical model with respect to the incumbent model parameter. Importantly, $\mathbf{V}_\pvec$ is also a function of $\xvec$ and $\pvec$ since it comprises an orthogonal basis for the row space of $\partial_\pvec f(\xvec,\pvec)$.

\section{\textbf{Latent state reconstruction}}  \label{sec:data_recnstrctn}

With a mean-field approximation, and assuming the spectroscopic data includes the dynamics of all kinetically relevant reaction intermediates in an NPT-system, calibration coefficients can be retrieved by minimizing the raw signal projection onto the coupling matrix nullspace. With the nomenclature in \cref{main:eq:res_interp}, i.e., taking $\tilde{\xvec}$ as observed values, the underlying material balance can be expressed in terms of the nullspace equations as in \cref{eq:null_spec}.
\begin{equation}
    \begin{aligned}
        (\mathbf{U}_\mathbf{o}^\sN)^T\left(\tilde{\xvec}_{\mathbf{o}(t_i)}-{\tilde{\mathbf{x}}}_{\mathbf{o}(t_j)}\right)+(\mathbf{U}_{\boldsymbol{\ast}}^\sN)^T\left(\tilde{\xvec}_{\boldsymbol{\ast}(t_i)}-{\tilde{\mathbf{x}}}_{\boldsymbol{\ast}(t_j)}\right)=\varepsilon_{\zvec_{ij}}^\sN
    \end{aligned}\label{eq:null_spec}
\end{equation}
In the absence of statistical noise, the nullspace error, $\varepsilon_{\zvec_i}^\sN,\,\forall{i}$, would be zero. Latent states can be represented in terms of raw signal, $\yvec\in\mathbb{R}_+^n$, and a calibration factor per chemical intermediate, $\boldsymbol{\gamma}\in\mathbb{R}_+^{\operatorname{dim}(\boldsymbol{\ast})}$, as in \cref{eq:null_calib}.
\begin{equation}
    \begin{aligned}
        \xvec_{\boldsymbol{\ast}}=\yvec_{\boldsymbol{\ast}}\circ\boldsymbol{\gamma}
    \end{aligned}\label{eq:null_calib}
\end{equation}
Coverage concentrations also need to be normalized, which can be expressed in terms of a parameterized set of solutions for $\boldsymbol{\gamma}$, as in the following minimization problem, \cref{eq:opt_norm}.
\begin{equation}
\begin{aligned}
    \boldsymbol{\gamma}(\boldsymbol{\beta})=\boldsymbol{\gamma}^\sR+\boldsymbol{\gamma}^\sN(\boldsymbol{\beta})=\left(\sum_{i}^n\tilde{\mathbf{y}}_{\boldsymbol{\ast}(t_i)}\tilde{\mathbf{y}}_{\boldsymbol{\ast}(t_i)}^T\right)^+\sum_{i}^n\tilde{\mathbf{y}}_{\boldsymbol{\ast}(t_i)}+{\mathbf{U}^\sN_{\boldsymbol{\gamma}}}\boldsymbol{\beta}
    \label{eq:opt_norm}
\end{aligned}    
\end{equation}
Where ${\mathbf{U}^\sN_{\boldsymbol{\gamma}}}$ is a basis for the nullspace of $\sum_{i}^n\tilde{\mathbf{y}}_{\boldsymbol{\ast}(t_i)}{\tilde{\mathbf{y}}_{\boldsymbol{\ast}(t_i)}}^T$, which can be computed from its eigendecomposition. By replacing \cref{eq:null_calib} with the partial solution of \cref{eq:opt_norm} in \cref{eq:null_spec}, the reduced form in \cref{eq:null_y} is obtained. For realistic scenarios, statistical noise will lead to non-zero eigenvalues, which are associated with the nullspace of $\sum_{i}^n\tilde{\mathbf{y}}_{\boldsymbol{\ast}(t_i)}{\tilde{\mathbf{y}}_{\boldsymbol{\ast}(t_i)}}^T$ and, therefore, a cutoff criterion must be established to obtain the nullspace basis by selecting the associated eigenvectors.
\begin{equation}
    \begin{aligned}
        &(\mathbf{U}_\mathbf{o}^\sN)^T\left(\tilde{\xvec}_{\mathbf{o}(t_i)}-{\tilde{\mathbf{x}}}_{\mathbf{o}(t_j)}\right)+(\mathbf{U}_{\boldsymbol{\ast}}^\sN)^T\operatorname{diag}\left(\tilde{\mathbf{y}}_{\boldsymbol{\ast}(t_i)}-{\tilde{\mathbf{y}}}_{\boldsymbol{\ast}(t_j)}\right)\boldsymbol{\gamma}(\boldsymbol{\beta})\\
        &=\mathbf{v}_{\mathbf{o}_{ij}}+\mathbf{V}_{\boldsymbol{\ast}_{ij}}\boldsymbol{\gamma}(\boldsymbol{\beta})
    \end{aligned}\label{eq:null_y}
\end{equation}
The minimization of the MSE of \cref{eq:null_y} in terms of $\boldsymbol{\beta}$, considering only the asymmetrical pairs $(i,j)\,:\,j>i$, leads to the closed-form solution in \cref{eq:null_min}:
\begin{equation}
    \begin{aligned}
        \boldsymbol{\gamma}=-{\mathbf{U}^\sN_{\boldsymbol{\gamma}}}\left((\mathbf{U}^\sN_{\boldsymbol{\gamma}})^T\left(\sum_i^n\sum_j^{i-1}\mathbf{V}_{\boldsymbol{\ast}_{ij}}^T\mathbf{V}_{\boldsymbol{\ast}_{ij}}\right)\mathbf{U}^\sN_{\boldsymbol{\gamma}}\right)^+(\mathbf{U}^\sN_{\boldsymbol{\gamma}})^T\sum_i^n\sum_j^{i-1}\mathbf{V}_{\boldsymbol{\ast}_{ij}}^T\left(\mathbf{v}_{\mathbf{o}_{ij}}+\mathbf{V}_{\boldsymbol{\ast}_{ij}}\boldsymbol{\gamma}^\sR\right)+\boldsymbol{\gamma}^\sR
    \end{aligned}\label{eq:null_min}
\end{equation}
With $\boldsymbol{\beta}$ from \cref{eq:null_min} replaced in \cref{eq:opt_norm}, multiplicative or calibration factors $\boldsymbol{\gamma}$ can be found to satisfy the chemical system nullspace condition in \cref{main:eq:cnull} under coverage normalization. 
~\begin{figure}[hbt!]
    \centering
    \includegraphics[keepaspectratio=true,scale=1.,clip=True]{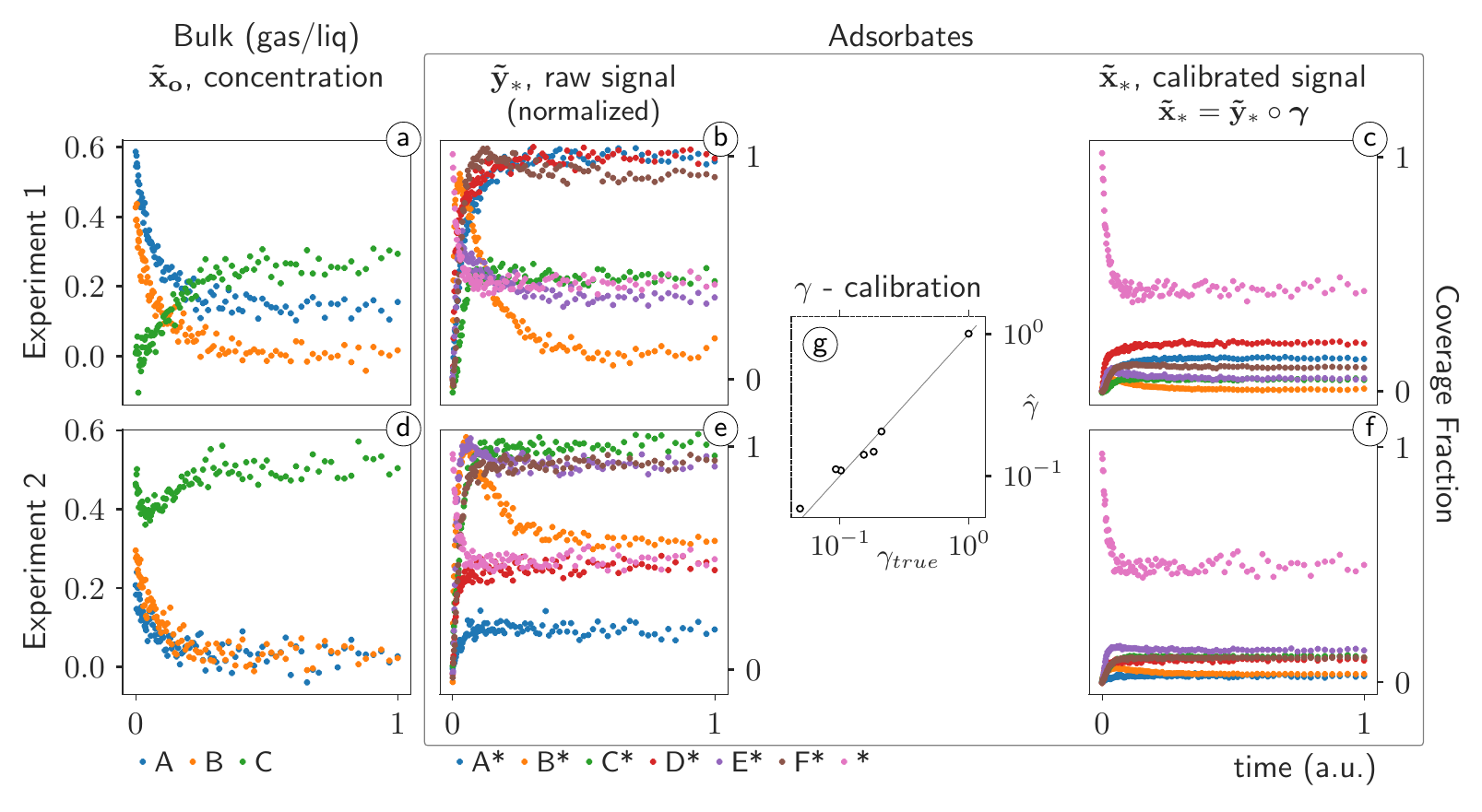}
    \caption{Synthetic experiments: $A$, $B$ and $C$ bulk phase concentrations (a and d); raw adsorbate signals (b and e), regressed calibration coefficients parity plot ($\boldsymbol{\hat{\gamma}}$) (g), and reconstructed/calibrated adsorbates coverage fractions (c and f), for \ic 1, $\{x_A,x_B,x_C,x_*\}_{t_0}=\{0.6,0.4,0.0,1.0\}$, and \ic 2, $\{x_A,x_B,x_C,x_*\}_{t_0}=\{0.2,0.3,0.5,1.0\}$.}\label{fig:calibration}
\end{figure}~

\section{Model Parameters Conditional Covariance - Calibration Factor}\label{sec:conditional_covar}

The calibration factors, $\boldsymbol{\gamma}$, can be included in the r\kinn{s} objective function by expressing them error terms for interpolation and model in the original space, \cref{main:eq:res_interp,main:eq:res_model}. We need first note that for reconstructed/calibrated data, the SAs have predefined nullspace invariances, $\zvec^\sN$, which is estimated directly from the unscaled data, $\tilde\yvec$, and optimal calibration factors, $\boldsymbol{\gamma}^*$, \cref{eq:null_min}, such that,
\begin{equation}
\begin{aligned}
    \xvec(t_i,\parsm,\boldsymbol{\gamma})&=\mathbf{U}^\sR\zvec^\sR(t_i,\parsm)+\mathbf{U}^\sN\zvec^\sN(\boldsymbol{\gamma})\\
    &=\mathbf{U}^\sR\zvec^\sR(t_i,\parsm)+\mathbf{U}^\sN\zvec^\sN\\
    &=\mathbf{U}^\sR\zvec^\sR(t_i,\parsm)+\mathbf{U}^\sN\frac{1}{n}\sum_{j=1}^n\left((\mathbf{U}^\sN)^T\tilde{\xvec}_j\odot\boldsymbol{\gamma}\right)\\
    &=\mathbf{U}^\sR\zvec^\sR(t_i,\parsm)+\frac{1}{n}\sum_{j=1}^n\left(\mathbf{U}^\sN(\mathbf{U}^\sN)^T\operatorname{diag}(\tilde{\xvec}_j)\right)\boldsymbol{\gamma}
\end{aligned}    
\end{equation}
Such a formulation conveys the effect of $\boldsymbol{\gamma}$ in the nullspace solution, i.e., linear shifts in the data and, consequently, on the SAs. However, to allow the propagation of errors to the range equations, the error expression need to be formulated based on the unscaled mapping $\yvec=\xvec\odot(\boldsymbol{\gamma}^*)^{-1}$.
\begin{equation}
\begin{aligned}
    \erresp_{\xvec_i}(\parsm,\boldsymbol{\gamma}) &= \xvec(t_i,\parsm,\boldsymbol{\gamma})-\tilde{\yvec}_i\odot\boldsymbol{\gamma}\\
    \therefore\erresp_{\yvec_i}=\erresp_{\xvec_i}(\parsm,\boldsymbol{\gamma})\odot(\boldsymbol{\gamma}^*)^{-1} &= \xvec(t_i,\parsm,\boldsymbol{\gamma})\odot(\boldsymbol{\gamma}^*)^{-1}-\tilde{\yvec}_i
\end{aligned}\label{eq:obs_res_gamma}
\end{equation}
Let $\yvec(t_i,\parsm,\boldsymbol{\gamma})=\xvec(t_i,\parsm,\boldsymbol{\gamma})\odot(\boldsymbol{\gamma}^*)^{-1}$, $\erresp_{\xvec_i}$ can be expressed in terms of $\yvec(t_i,\parsm,\boldsymbol{\gamma})$ as follows, 
\begin{equation}
\begin{aligned}
\erresp_{\xvec_i} &= \yvec(t_i,\parsm,\boldsymbol{\gamma})\odot\boldsymbol{\gamma}-\tilde{\yvec}_i\\
    \erresp_{\dxvec_i}&=\dot{\xvec}(t_i,\parsm,\boldsymbol{\gamma})- \mathbf{M}\left(\mathbf{k}(\theta,\pvec)\circ \fmap(\xvec(t_i,\parsm))\right)\\
    &=\dot{\yvec}(t_i,\parsm,\boldsymbol{\gamma})\odot\boldsymbol{\gamma}- \mathbf{M}\left(\mathbf{k}(\theta,\pvec)\circ \fmap(\dot{\yvec}(t_i,\parsm,\boldsymbol{\gamma})\odot\boldsymbol{\gamma})\right)\label{eq:ode_nn_mkm_gamma}
\end{aligned}
\end{equation}
With $\erresp_{\zvec_i}^{\sR/\sN}=(\mathbf{U}^{\sR/\sN})^T\erresp_{\xvec_i}$ and $\erresp_{\dzvec_i}^{\sR/\sN}=(\mathbf{U}^{\sR/\sN})^T\erresp_{\dxvec_i}$, the negative log-likelihood after range-nullspace decomposition can be evaluated as in \cref{main:eq:loglikhd_rn}, and the Fisher information matrix, $\mathcal{I}$, which is equivalent to the Hessian matrix of the MLE, $\lt$, can be estimated as a function of $\pvec$ and $\boldsymbol{\gamma}$.
\begin{equation}
    \begin{aligned}
        \mathcal{I}(\pvec,\boldsymbol{\gamma})=\nabla^2_{[\pvec\;\boldsymbol{\gamma}]}\lt&=\begin{bmatrix}\partial^2_\pvec\lt&\partial_{\pvec} \partial_{\boldsymbol{\gamma}}\lt\\\partial_{\boldsymbol{\gamma}} \partial_\pvec\lt&\partial^2_{\boldsymbol{\gamma}}\lt\end{bmatrix}
    \end{aligned}
\end{equation}
The corresponding asymptotic covariance matrix if given by:
\begin{equation}
    \begin{aligned}
        \boldsymbol{\Sigma}_{[\pvec\;\boldsymbol{\gamma}]}=\frac{1}{n}(\mathcal{I}(\pvec,\boldsymbol{\gamma}))^{-1}=\begin{bmatrix}\boldsymbol{\Sigma}_{\pvec\pvec}&\boldsymbol{\Sigma}_{\pvec\boldsymbol{\gamma}}\\\boldsymbol{\Sigma}_{\boldsymbol{\gamma}\pvec}&\boldsymbol{\Sigma}_{\boldsymbol{\gamma}\boldsymbol{\gamma}}\end{bmatrix}
    \end{aligned}
\end{equation}
Where the diagonal terms represent the covariance matrices for model parameters and calibration factors, and off-diaconal terms are the covariances between them. The conditional covariance for the kinetic model parameters, $\boldsymbol{\Sigma}_\pvec$ is evaluated as:
\begin{equation}
    \boldsymbol{\Sigma}_\pvec = \boldsymbol{\Sigma}_{\pvec\pvec}+\boldsymbol{\Sigma}_{\pvec\boldsymbol{\gamma}}\left(\boldsymbol{\Sigma}_{\boldsymbol{\gamma}\boldsymbol{\gamma}}\right)^{-1}\boldsymbol{\Sigma}_{\boldsymbol{\gamma}\pvec}
\end{equation}
Which can be from Taylor expansion of the residuals assuming $\boldsymbol{\gamma}$ and $\pvec$ uncorrelated and normally distributed, and using Cholesky decomposition of the covariance matrices to generated perturbations around the optimal values.

\section{Reaction Network Type \textit{dcs}}\label{sec:structure}

The shape of the stoichiometry matrices reflect the elementary reactions of the system such that the rows are associated with the mass balance of each species, and the columns are associated with reaction rates, therefore having the same size as kinetic parameters k. The \textit{s}-type elementary steps embody elementary reactions between surface intermediates exclusively. In the following, $D*$, $E*$ and $F*$ are reaction intermediates for which there are no stable desorbed counterparts, which cannot be directly measured or inferred from ordinary analytical chemistry techniques.

\begin{align}
    A+*&\underset{k_{\text{-}1}^d}{\stackrel{k_1^d}{\rightleftharpoons}} A*\label{reac:adsA}\tag{\textbf{d}.1}\\
    B+*&\underset{k_{\text{-}2}^d}{\stackrel{k_2^d}{\rightleftharpoons}} B*\label{reac:adsB}\tag{\textbf{d}.2}\\
    C+*&\underset{k_{\text{-}3}^d}{\stackrel{k_3^d}{\rightleftharpoons}} C*\label{reac:adsC}\tag{\textbf{d}.3}\\
    A*+*&\underset{k_{\text{-}3}^c}{\stackrel{k_3^c}{\rightleftharpoons}} 2D*\tag{\textbf{c}.1}\\
    B*+*&\underset{k_{\text{-}4}^c}{\stackrel{k_4^c}{\rightleftharpoons}} 2E*\tag{\textbf{c}.2}\\
    D*\:+\:E*&\underset{k_{\text{-}1}^s}{\stackrel{k_{1}^s}{\rightleftharpoons}} F*\:+\:*\tag{\textbf{s}.1}\\
    F*\:+\:E*&\underset{k_{\text{-}5}^c}{\stackrel{k_5^c}{\rightleftharpoons}} C*\:+\:*\tag{\textbf{c}.3}
\end{align}

\begin{equation}
\begin{aligned}
    \ln(\mathbf{k}_0)^T=\left[3.00\:\;2.08\:\;3.18\:\;2.48\:\;2.77\:\;3.69\:\;6.46\:\;6.87\:\;5.08\:\;4.38\:\;6.46\:\;5.48\:\;6.33\:\;5.08\right]
    \\\\
    M=
    \left[
    \begin{array}{rrrrrrrrrrrrrr}
     -1&1&0&0&0&0&0&0&0&0&0&0&0&0 \\
      0&0&-1&1&0&0&0&0&0&0&0&0&0&0 \\
      0&0&0&0&-1&1&0&0&0&0&0&0&0&0 \\
      1&-1&0&0&0&0&-1&1&0&0&0&0&0&0 \\
      0&0&1&-1&0&0&0&0&-1&1&0&0&0&0 \\
      0&0&0&0&1&-1&0&0&0&0&0&0&1&-1 \\
      0&0&0&0&0&0&2&-2&0&0&-1&1&0&0 \\
      0&0&0&0&0&0&0&0&2&-2&-1&1&-1&1 \\
      0&0&0&0&0&0&0&0&0&0&1&-1&-1&1 \\
     -1&1&-1&1&-1&1&-1&1&-1&1&1&-1&1&-1
    \end{array}
    \right]\\\\
    \mathbf{x}^T=\left[x_{A}\;\;x_{B}\;\;x_{C}\;\;x_{A*}\;\;x_{B*}\;\;x_{C*}\;\;x_{D*}\;\;x_{E*}\;\;x_{F*}\;\;x_{*}\right]
\end{aligned}
\end{equation}

%\clearpage

\bibliography{supp}
\bibliographystyle{unsrt} % GSG change unsrt